%% file: iclr2026_conference.tex
\documentclass{article} 

\usepackage{iclr2026_conference,times}

\input{math_commands.tex}

\usepackage{url}
\usepackage{booktabs}
\usepackage{xcolor}         
\definecolor{cvprblue}{rgb}{0.21,0.49,0.74}
\usepackage{multirow}
\usepackage{graphicx}

\usepackage[pagebackref,breaklinks,colorlinks,citecolor=cvprblue,urlcolor=cvprblue,linkcolor=cvprblue]{hyperref}
\usepackage{caption}
\usepackage{pifont}
\usepackage{colortbl}
\usepackage{wrapfig}
\usepackage{ulem}
\usepackage{amsthm}
\usepackage{longtable}
\usepackage{enumitem}

\title{Scalable Multilingual Multimodal Machine Translation with Speech-Text Fusion}

\author{
    \begin{tabular}{c}
    \\
        \textbf{Yexing Du}$^{1,2}$, \textbf{Youcheng Pan}$^{2}$\thanks{Corresponding Authors: \texttt{\{panych,xiangy\}@pcl.ac.cn}, \texttt{mliu@ir.hit.edu.cn}}~, \textbf{Zekun Wang}$^{1}$, \textbf{Zheng Chu}$^{1}$, \textbf{Yichong Huang}$^{1}$ \\
        \textbf{Kaiyuan Liu}$^{1,2}$, \textbf{Bo Yang}$^{2}$, \textbf{Yang Xiang}$^{2 }$\footnotemark[1]~, \textbf{Ming Liu}$^{1,2}$\footnotemark[1]~, \textbf{Bing Qin}$^{1,2}$ \\\\
       \small{\textnormal{$^1$Harbin Institute of Technology \quad $^2$Pengcheng Laboratory}}
    \end{tabular}
}

\iclrfinalcopy
\begin{document}

\maketitle

\begin{abstract}
Multimodal Large Language Models (MLLMs) have achieved notable success in enhancing translation performance by integrating multimodal information.
However, existing research primarily focuses on image-guided methods, whose applicability is constrained by the scarcity of multilingual image-text pairs.
The speech modality overcomes this limitation due to its natural alignment with text and the abundance of existing speech datasets, which enable scalable language coverage.
In this paper, we propose a \textbf{Speech-guided Machine Translation (SMT)} framework that integrates speech and text as fused inputs into an MLLM to improve translation quality.
To mitigate reliance on low-resource data, we introduce a \textbf{Self-Evolution Mechanism}.
The core components of this framework include a text-to-speech model, responsible for generating synthetic speech, and an MLLM capable of classifying synthetic speech samples and iteratively optimizing itself using positive samples.
Experimental results demonstrate that our framework surpasses all existing methods on the Multi30K multimodal machine translation benchmark, achieving new state-of-the-art results.
Furthermore, on general machine translation datasets, particularly the FLORES-200, it achieves average state-of-the-art performance in 108 translation directions. Ablation studies on CoVoST-2 confirms that differences between synthetic and authentic speech have negligible impact on translation quality. The code and models are released at \url{https://github.com/yxduir/LLM-SRT}.
\end{abstract}

\section{Introduction}

Multimodal Machine Translation (MMT) leverages complementary information from multiple modalities, such as images, to enhance machine translation (MT) quality. These modalities provide supplementary contextual information for source texts, thereby mitigating ambiguities caused by polysemy or omissions~\citep{shen2024survey}.

\begin{figure}[t]
    \centering
     
    \begin{minipage}[t]{0.3\textwidth}
        \centering
        
        \includegraphics[width=\linewidth]{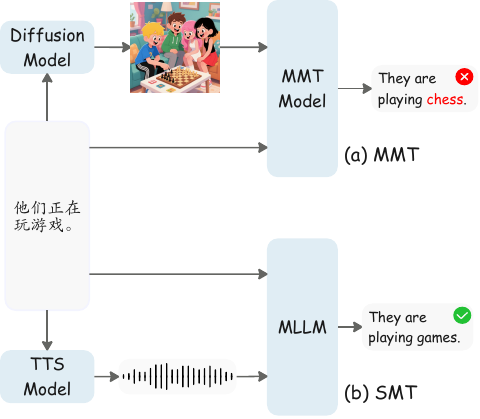} 
        \vspace{0.68cm} 
        \caption{Image-Guided vs. Speech-Guided Machine Translation.}
        \label{fig:intro}
        
    \end{minipage}
    \hfill
    \begin{minipage}[t]{0.68\textwidth}
        \centering
        \tiny
        \renewcommand{\arraystretch}{1.175} 
        \setlength{\tabcolsep}{1pt} 
         \vspace{-3.5cm} 
        \resizebox{\textwidth}{!}{
        \begin{tabular}{l c|l c}
    \toprule
    \textbf{Image Dataset} & \textbf{Language} & \textbf{Speech Dataset}& \textbf{Language} \\ 
    \midrule
    IAPR TC-12~\citep{Grubinger_2006} & \multicolumn{1}{c|}{eng, deu} &MuST-C~\citep{di2019must}& eng$\rightarrow$15\\ 
    Multi30K~\citep{DBLP:conf/acl/ElliottFSS16} & \multicolumn{1}{c|}{eng, deu, deu, fra}&CoVoST-2~\citep{wang2020covost}& 21$\rightarrow$eng eng$\rightarrow$15\\  
    
    {MLT}~\citep{DBLP:conf/lrec/LalaS18} &  \multicolumn{1}{c|}{eng, deu}&Europarl-ST~\citep{iranzo2020europarl}& 9$\leftrightarrow$9 \\ 
    
    MultiSense~\citep{DBLP:conf/naacl/GellaEK19}  & \multicolumn{1}{c|}{eng, deu, spa}&Fleurs~\citep{fleurs2022arxiv}& 102$\leftrightarrow$102 \\  
    
    AmbigCaps~\citep{li2021vision} & \multicolumn{1}{c|}{eng, tur}&Granary~\citep{koluguri2025granary}& 25$\rightarrow$eng\\  
    
    Fashion-MMT~\citep{DBLP:conf/mm/0003CJLXH21} & \multicolumn{1}{c|}{eng, cmn} &CCFQA~\citep{du2025ccfqa}&8$\leftrightarrow$8\\ 
   
    EMMT~\citep{DBLP:conf/acl/ZhuSCHWW23} & \multicolumn{1}{c|}{eng, cmn} &BhasaAnuvaad~\citep{sankar-etal-2025-towards}&eng$\leftrightarrow$14 \\  
    TIT Dataset~\citep{DBLP:conf/icpr/MaZTHWZ022} & eng, deu, cmn \\
    BLATID~\citep{DBLP:journals/tmm/ChenYYL23} & eng, cmn \\  
    OCRMT30K~\citep{DBLP:conf/acl/LanYLZ0WHS23} & eng, cmn \\ 
    MSCTD~\citep{DBLP:conf/acl/LiangMXCZ22} & eng, deu, cmn \\ 
    BIG-C~\citep{DBLP:conf/acl/SikasoteMAA23} & eng, ben \\ 
    HaVQA~\citep{DBLP:conf/acl/ParidaAMBKAKSBK23} & eng, hau \\ 

    \multirow{1}{*}{$\mathrm{M}^{3}$~\citep{DBLP:conf/emnlp/GuoLHYL0C22}} & \multicolumn{1}{c|}{eng$\rightarrow$6}\\  

    \bottomrule
    \end{tabular}}
        \captionof{table}{Dataset Statistics. For the languages supported by the image datasets, please refer to Table~\ref{tab:mmt_langs}. Our MLLM supports 28 languages, as shown in Table~\ref{tab:langs}.}
        \label{tab:example}
    \end{minipage}
\end{figure}

Traditionally, image-based MMT models~\citep{cheng-etal-2024-soul} process image-text pairs to generate translations, leveraging visual context for semantic disambiguation. However, these models require an associated image for each input text, which limits their applicability. 
Recent image-free approaches~\citep{guo2023bridging} have employed diffusion models~\citep{rombach2022high} to generate synthetic images to enhance translation. While these studies  address the issue of image dependency, those methods still face two limitations:
(1) \textbf{Generalizability}: While MMT models perform well on ambiguous datasets~\citep{DBLP:conf/acl/ElliottFSS16}, they struggle to generalize to general translation datasets and even introduce noise in some scenarios (see Figure~\ref{fig:intro}).
(2) \textbf{Multilinguality}: Existing image MMT datasets~\citep{DBLP:conf/emnlp/GuoLHYL0C22} support only a few languages, with limited of languages coverage (see Table~\ref{tab:example}). Advances in diffusion Text-to-Speech (TTS) models~\citep{du2024cosyvoice} have achieved high-quality, zero-shot multilingual speech synthesis. This raises a question: \textbf{Can we leverage speech modalities to enhance translation quality}?

Recent studies have revealed that, alongside lexical information, speech signals also convey prosodic cues, which offer valuable supplementary information~\citep{chi2025role}. Inspired by fusion of text and prosody features, we propose the framework of Speech-guided Machine Translation (SMT), which maps speech-text fusion inputs \{\textit{speech}, \textit{text}\} to \{\textit{translation}\} outputs.
Specifically, our SMT framework integrates a TTS model with an MLLM through a self-evolution mechanism~\citep{tao2024survey} that leverages synthetic speech to enhance translation performance.

The framework consists of two core components: 
(1) \textbf{MLLM Pre-training}: We employ a multi-stage curriculum learning strategy with progressively complex objectives, beginning with speech recognition (ASR) for speech-text mapping, then speech-to-text translation (S2TT) for cross-lingual and cross-modality bridging, and culminating in SMT training for joint speech-text processing.
(2) \textbf{Self-Evolution Mechanism}: This component synthesizes training data via the TTS model, where the MLLM classifies speech samples based on translation scores. The MLLM undergoes continuous training using positive samples, while translation performance metrics serve as evolution objectives, enabling continuous framework improvement through iterative refinement cycles.

The experimental results demonstrate that our framework achieves new state-of-the-art (SOTA) results on the Multi30K benchmark~\citep{DBLP:conf/acl/ElliottFSS16}, surpassing all existing MMT approaches. Our framework further achieves SOTA average machine translation (MT) performance across 108 languages directions on the FLORES-200 benchmark~\citep{nllb2022}, outperforming much larger language models. Ablation studies on the CoVoST-2 dataset~\citep{wang2020covost} also reveal that the discrepancy between synthetic and authentic speech has a negligible effect on translation performance. In summary, our key contributions are as follows:

\label{contributions}
\begin{itemize}[itemsep=3pt, topsep=3pt, parsep=0pt]
    \item We propose a novel speech-guided machine translation framework, which consists of a TTS model and an MLLM. Our framework leverages prosodic cues in speech to enhance translation performance and supports 28 languages.

    \item We propose a self-evolution framework that autonomously generates training data for iterative self-enhancement. The framework employs continual training for the MLLM, utilizing synthetic data to improve the model's low-resource translation quality.
    
    \item Our framework achieves state-of-the-art results on MMT and MT tasks across multiple benchmarks (Multi30K, FLORES-200). Ablation studies on the CoVoST-2 benchmark show that the difference between authentic and synthetic speech has a negligible impact on translation performance.
\end{itemize}

\section{Methodology}
\subsection{Modality-Agnostic Hypothesis} \label{subsec:hypothesis}
\paragraph{Assumption 1.}
Any auxiliary modality can enhance machine translation performance when:
\begin{itemize}[itemsep=2pt, topsep=2pt, parsep=0pt]
    \item The modality provides semantically relevant information to the source text.
    \item The modality representation can be aligned and jointly optimized with textual features in a shared latent space, given sufficient training data to learn discriminative embeddings.
\end{itemize}

\begin{figure*}[t] \centering
\includegraphics[width=\textwidth]{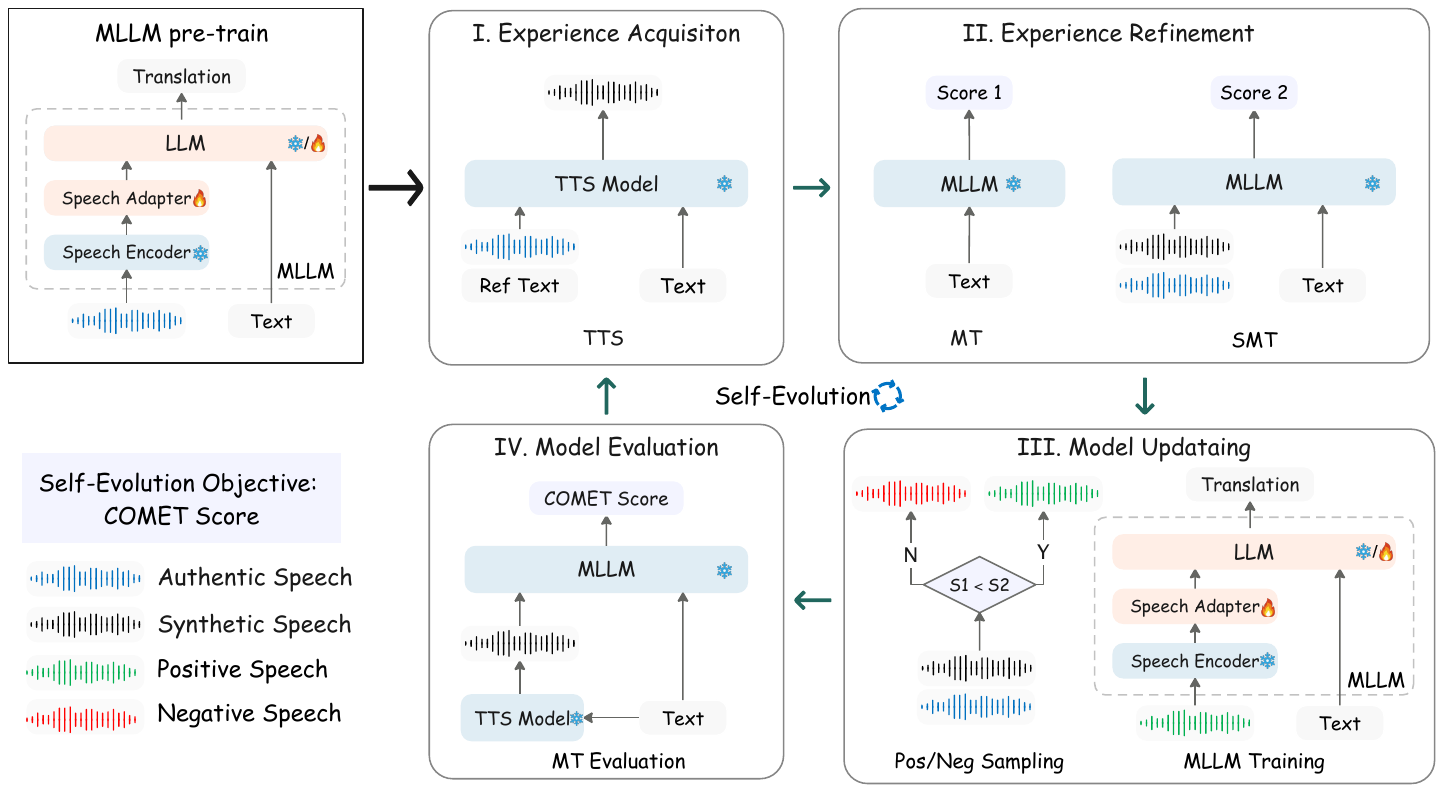} 
\caption{Overview of Our SMT Framework. The proposed system architecture comprises two core components: (1) MLLM pretraining and (2) Self-Evolution. This framework takes text input, synthesizes speech of the text via a TTS model, and leverages the MLLM to process both text and speech features for higher-quality translation output. Self-evolution mechanism can autonomously generate training data to iteratively optimize the framework.} 
\label{framework}
\end{figure*}

\subsection{Overall Design}
Figure \ref{framework} illustrates the SMT framework, comprising an MLLM and a TTS model. The processing pipeline operates as follows: First, the system accepts textual input and synthesizes speech via the TTS model. Then, the MLLM processes both the text and synthetic speech to generate translations. The following subsections detail two key components: MLLM pretraining (Section \ref{sec:mllm}) and self-evolution mechanism (Section \ref{aligment_method}).

\subsection{MLLM Pre-training}
\label{sec:mllm}

The MLLM is built upon a large language model (LLM)~\citep{cui2025multilingual}, adopts Whisper's encoder~\citep{radford2023robust} as the speech encoder, followed by a Q-Former~\citep{li2023blip} and MLP layer for speech adapter. We design a three-stage training pipeline and perform instruction tuning. The sequential fine-tuning stages comprise: (1) automatic speech recognition, (2) speech-to-text translation, and (3) speech-guided machine translation.

\begin{wraptable}[12]{r}{0.57\textwidth}
    \centering
    \footnotesize
    \vspace{-0.3cm}
    \label{Parameters}
            \renewcommand{\arraystretch}{1} 
    \resizebox{0.58\textwidth}{!}{
    \setlength{\tabcolsep}{2pt}{
    \begin{tabular}{lrcl}
        \toprule
        \multirow{2}{*}{\textbf{Modules}}&\multirow{2}{*}{\textbf{Param}}&\multirow{2}{*}{\textbf{Stage}}&\multirow{2}{*}{\textbf{Details}}  \\
        &&& \\
        \midrule 
        Speech Encoder&$\sim$635M&-&Whisper's encoder  \\
        \midrule
        Speech Adapter&{\color{cvprblue}$\sim$80.5M}&{All}& Q-Former and MLP \\
        \midrule
        LLM& $\sim$9.2B&-&GemmaX2-28-9B \\
        \midrule
        LLM adapter&{\color{cvprblue}$\sim$8.9M}& III& LoRA (r=16, alpha=32) \\
        \midrule
        Total&$\sim$10B&\\
        \bottomrule
    \end{tabular}
    }
    }
\caption{MLLM Pre-training. The {\color{cvprblue}blue color} indicates the number of trainable parameters. }
\end{wraptable}

\paragraph{ASR.} 
The MLLM learns speech-text alignment through ASR pre-training while keeping only the speech adapter trainable. 

\paragraph{S2TT.} 
Given speech input and instructions, the MLLM simultaneously generates transcriptions and translations.

\paragraph{SMT.} 
The MLLM processes joint speech-text inputs to generate translation outputs by leveraging complementary multimodal information.

\newpage

\subsection{Self-Evolution Mechanism}
\label{aligment_method}
Self-evolution mechanism allows models to autonomously learn through four phases: experience acquisition, experience refinement, updating, and evaluation. Our SMT framework is
based on (1) MLLM, (2) TTS model, and (3) a S2TT dataset with authentic speech, text, and translation.

\subsubsection{Stage I: Experience Acquisition}
The purpose of this stage is to generate synthetic speech.
During this stage, the prompt text and the predicted speech duration are strictly aligned with authentic speech and text pairs.

\paragraph{TTS Inference.}
We employ a TTS model to synthesize speech signals from the text in the S2TT dataset. Given a reference text, the TTS model generates a new speech utterance while cloning a randomly selected voice from the same dataset. This process ensures a diverse set of synthetic speech data with varied prosody, which is crucial for our framework's training.

\subsubsection{Stage II: Experience Refinement} 
This stage implements a quality-aware labeling strategy for speech samples. We find that not all speech is beneficial for translation, so we need to classify the samples. This process is achieved by comparing the scores of MT and SMT.

\paragraph{MT and SMT Inference.}
The MLLM operates in two distinct modes. In MT mode, the model processes textual inputs $t_{\text{text}}$ to generate translations $t_{\text{trans}}$, producing score $S_1$. In SMT mode, the model accepts either authentic speech $s_{\mathrm{ref}}$ or synthetic speech $s_{\mathrm{gen}}$ paired with its corresponding text input to generate translations, producing score $S_2$.

\subsubsection{Stage III: Model Updating}
This stage is dedicated to optimizing the MLLM by leveraging the synthetic data generated in the previous stage. The primary goal is to enhance the MLLM's ability to effectively utilize prosodic cues from speech input for improved translation quality.

\paragraph{Positive/Negative Sampling.}
We first perform a comparative analysis to categorize each synthesized speech-text pair into either a positive ($s_{\text{pos}}$) or a negative ($s_{\text{neg}}$) sample. Let $S_1$ be the translation quality score with text input only, and $S_2$ be the score when the MLLM receives both text and speech input.

A sample is categorized as a \textbf{positive sample} ($s_{\text{pos}}$) if the additional speech input improves translation performance ($S_2 > S_1$). Conversely, a sample is labeled as a negative sample ($s_{\text{neg}}$) if the speech input provides no benefit ($S_2 \leq S_1$). The scores are computed as:
\begin{equation}
    \begin{cases}
        S_1 = \text{COMET}\Big(\text{MLLM}\big(t_{\text{text}}\big),\, t_{\text{trans}}\Big) \\[10pt]
        S_2 = \text{COMET}\Big(\text{MLLM}\big( s_{\mathrm{ref}} \text{ or } s_{\mathrm{gen}},t_{\text{text}}\big),\, t_{\text{trans}}\Big)
    \end{cases}
\end{equation}

\paragraph{MLLM Continuous Training.}
The MLLM is then continually fine-tuned using only the identified positive samples ($s_{\text{pos}}$). This targeted training strategy guides the model to prioritize and learn from the most beneficial speech-text interactions, thereby enhancing its ability to leverage prosody for superior translation performance.

\subsubsection{Stage IV: Model Evaluation}
In this final stage, we evaluate the framework's translation performance to determine whether to continue the self-evolution loop. We synthesize speech for the evaluation text using a fixed reference voice and measure the SMT framework's performance with the COMET score. This process iterates until the COMET score on the evaluation set converges and no longer shows significant improvement.

\newpage

\section{Experiments}
\label{sec:experiment setup}

\subsection{Datasets}
We conduct comprehensive evaluations on several benchmarks. For multimodal machine translation, we use Multi30K\footnote{\url{https://github.com/multi30k/dataset}}~\citep{DBLP:conf/acl/ElliottFSS16}. For machine translation, we use FLORES-200\footnote{\url{https://github.com/facebookresearch/flores}}~\citep{nllb2022} and {WMT24++}\footnote{\url{https://huggingface.co/datasets/google/wmt24pp}}~\citep{deutsch2025wmt24++}. Additionally, we perform ablation studies on the CoVoST-2\footnote{\url{https://github.com/facebookresearch/covost}} dataset~\citep{wang2020covost}. Detailed information for datasets is provided in Table \ref{tab:benchmark}.

\subsection{Experiment Setup}
\paragraph{Model Architecture.} Our MLLM consists of a frozen speech encoder, specifically the encoder from Whisper-large-v3~\citep{radford2023robust}, and a trainable adapter layer. This adapter comprises a Q-Former~\citep{DBLP:conf/icml/0008LSH23} and a multilayer perceptron (MLP). The LLM backbone is GemmaX2-28-9B~\citep{cui2025multilingual}. Following the configuration in~\citep{yu2024connecting}, our Q-Former uses 80 queries, each with a dimension of 768. The datasets used for MLLM training are detailed in Table \ref{tab:train}. For the TTS model, we adopt the CosyVoice2~\citep{du2024cosyvoice} model.

\paragraph{Training Details.}

Experiments are conducted on four A100 GPUs (80GB). Following the experimental setup~\citep{ma2026slam}, we used the AdamW optimizer~\citep{adamw} with a peak learning rate of $1 \times 10^{-4}$. The learning rate was linearly warmed up over 1K steps and then linearly decayed for the remainder of the training. The models can be trained in under a week.

\paragraph{Evaluation Metrics.} For evaluation, we employ BLEU\footnote{\url{https://github.com/mjpost/sacrebleu}}~\citep{post-2018-call}, spBLEU~\citep{nllb2022}, and COMET\footnote{\url{https://huggingface.co/Unbabel/wmt22-comet-da}}~\citep{rei2020comet}. We compute spBLEU using the tokenizer "flores200". For a fair comparison, our LLM inference uses vLLM~\citep{kwon2023efficient}, with all beam search settings and temperature uniformly set to 1 and 0, respectively.

\subsection{Comparing Models}
\paragraph{MT Models.} We evaluate the translation performance of four models: Deepseek-V3.1 API~\citep{guo2025deepseek}, Gemma3-27B-it~\citep{team2025gemma}, Qwen3-Next-80B-A3B-Instruct~\citep{qwen3-blog-2024}, and NLLB-54B~\citep{nllb2022}.

\paragraph{MMT Models.} We compare our framework against two categories of existing multimodal machine translation models. We compare against four traditional MMT models that use text and authentic image: Soul-Mix~\citep{cheng-etal-2024-soul}, RG-MMT-EDC~\citep{DBLP:journals/talip/TayirL24}, WRA-guided~\citep{DBLP:journals/taslp/ZhaoKKC22}, and ConsQA-MMT~\citep{gao2025multimodal}. Additionally, we compare against four image-free MMT models that rely on text and synthetic image: VALHALLA~\citep{li2022valhalla}, Bridge~\citep{guo2023bridging}, DreamLLM~\citep{DBLP:conf/iclr/DongHPQGYZSZWK024}, and IMAGE~\citep{chen2024make}.

\subsection{Overall Results}
Our comprehensive experiments demonstrate the significant effectiveness of our proposed speech-guided machine translation approach. Our framework achieves new state-of-the-art results on the Multi30K benchmark, surpassing traditional text-only and image-based MMT models. SMT-9B also consistently outperforms much larger text-only language models. Furthermore, our framework shows strong generalization, achieving state-of-the-art results in 108 translation directions on the FLORES-200 benchmark. Finally, ablation studies confirm that the performance difference between authentic and synthetic speech is negligible.

\newpage
\subsubsection{Main Results for Multimodal Machine Translation}

\begin{table*}[t]\centering
    \small
    \setlength{\tabcolsep}{3pt} 
        \renewcommand{\arraystretch}{1.2}
    \vspace*{-0cm} 
    \resizebox{\textwidth}{!}{ 
        \begin{tabular}{l|ccc|ccc|cc}
            \toprule
             \multirow{2}{*}{\textbf{Models}}& & {eng $\rightarrow$ deu}&  & &{eng $\rightarrow$ fra}& & \multicolumn{2}{c}{{eng $\rightarrow$ ces}} \\ 
              & Test2016        & Test2017       & MSCOCO         & Test2016       & Test2017       & MSCOCO         & Test2016& Test2018\\ \midrule

            \rowcolor{gray!20} 
            \multicolumn{9}{c}{\textbf{Models based on Text}}\\ 
            \midrule
            
            DeepSeek-V3.1~\citep{guo2025deepseek}     &  44.2 / \uline{87.3} & 	\uline{41.1} / \uline{86.8} & 	36.4 / \uline{83.2} & 	55.3 / 88.2 	& 54.0 / 87.7 	& \uline{53.5} / \uline{85.8} 	&  \uline{37.9} / \uline{90.7}& \uline{35.9} / \uline{89.7}  \\ 
                       
             Gemma3-27B-it~\citep{team2025gemma}  &  43.7 / 87.1&	40.3 / 86.3&	36.1 / 83.2	&55.4 / 87.9	&54.3 / \uline{87.9}	&49.6 / 85.0	&36.4 / 89.9 &\uline{35.9} / 89.1 \\ 
             NLLB-moe-54B~\citep{nllb2022}  & 41.4 / 86.2&	39.7 / 85.8	&34.7 / 82.1	&55.1 / 87.4&	54.8 / 87.7	&53.3 / 85.3	&35.7 / 88.9 &  35.8 / 88.3 \\ 
                 Qwen3-Next-80B-A3B~\citep{qwen3-next-blog-2025} & 41.6 / 86.3 &	37.6 / 85.9 &	31.9 / 82.5 &	53.2 / 87.8 &	51.9 / 87.6 &	50.4 / 85.1 &	29.2 / 87.2 & 27.9 / 85.9 \\ \midrule

            \rowcolor{gray!20}  \multicolumn{9}{c}{\textbf{Models based on  Text  \& Authentic Image}}\\ \midrule  

            WRA-guided~$\dagger$~\citep{DBLP:journals/taslp/ZhaoKKC22}      & 39.3 / -----        & 32.3 / -----       & 28.5 / -----       & 61.8 / -----       & 54.1 / -----       & 43.4 / -----       & -----   &  -----   \\
            RG-MMT-EDC~$\dagger$~\citep{tayir2024encoder}      & 42.0 / -----        & 33.4 / -----       & 30.0 / -----       & 62.9 / -----       & 55.8 / -----       & 45.1 / -----       & -----   & -----    \\
            Soul-Mix~$\dagger$~\citep{cheng-etal-2024-soul}        & 44.2 / -----        & 37.1 / -----       & 34.2 / -----       & 64.7 / -----       & 57.4 / -----       & 49.2 / -----       & 36.5 / -----  &  32.8 / ----- \\
            ConsQA-MMT~$\dagger$~\citep{gao-etal-2025-multimodal} & 44.2 / -----  & 37.6 / -----  & 34.3 / -----  & 64.8 / -----  & 58.3 / -----  & 48.5 / -----   & 34.7 / ----- & 30.3 / ----- \\\midrule
            \rowcolor{gray!20} \multicolumn{9}{c}{\textbf{Models based on Text \& Synthetic Image}}\\ \midrule
            VALHALLA~$\dagger$~\citep{li2022valhalla}  & 42.7 / -----  & 35.1 / ----- & 30.7 / ----- & 63.1 / ----- & 56.0 / ----- & 46.5 / ----- &  -----   &  ----- \\
            Bridge~$\dagger$~\citep{guo2023bridging}        & 42.5 / -----  & 36.0 / ----- & 32.0 / ----- & 63.7 / ----- & 56.2 / ----- & 46.3 / ----- & 35.2 / ----- & 31.2 / -----\\
            DreamLLM~$\dagger$~\citep{DBLP:conf/iclr/DongHPQGYZSZWK024}       & 27.2 / 74.8  & 19.5 / 73.5 & 19.3 / 69.4 & 36.9 / 81.1 & 34.7 / 80.6 & 36.6 / 79.2 &  -----   &  ----- \\
            IMAGE~$\dagger$~\citep{chen-etal-2025-make}   & \uline{45.3} / 83.1  & 38.6 / 81.9 &  \uline{37.5} / 78.8 & \uline{67.5} / \uline{88.3} & \uline{61.5} / 86.6 & 49.3 / 82.5 & -----   &  ----- \\ \midrule
            \rowcolor{gray!20}   \multicolumn{9}{c}{\textbf{Models based on Text \& Synthetic Speech}}\\ \midrule
            {Baseline (Text only)}   & 42.9 / 87.0 & 38.8 / 86.4 & 34.3 / 82.7& 52.4 / 87.7 & 52.0 / 87.9 & {52.6} / 86.1 & 34.1 / 89.9 & 34.8 / 89.0\\
            {Baseline + Lora (Text only)}   & {44.0} / {87.0} & {39.4} / {86.4} & {35.3} / {83.0} & 55.5 / {88.1} & 54.0 / {88.2} & {53.4} / {85.9} & 37.2 / 90.0 & 35.7 / 89.1 \\
            SMT-9B   & \colorbox{cvprblue!14}{47.0}/\colorbox{cvprblue!14}{87.8}  & \colorbox{cvprblue!14}{41.8}/\colorbox{cvprblue!14}{87.3} & \colorbox{cvprblue!14}{38.5}/\colorbox{cvprblue!14}{84.0} & {67.0} /\colorbox{cvprblue!14}{90.0} & \colorbox{cvprblue!14}{62.1}/\colorbox{cvprblue!14}{89.6} & \colorbox{cvprblue!14}{55.3}/\colorbox{cvprblue!14}{86.7} &  \colorbox{cvprblue!14}{41.4} / \colorbox{cvprblue!14}{90.8}& \colorbox{cvprblue!14}{39.9} / \colorbox{cvprblue!14}{89.8} \\ 
            
            \bottomrule
        \end{tabular}
    }
    \raggedright{\scriptsize{\hspace*{0.5em}\underline{Underlined} denotes previous state-of-the-art models, while \colorbox{cvprblue!14}{highlighted} surpasses the previous models.}}
    \caption{Translation Performance on Multi30K (BLEU / COMET) MMT Benchmark. {The average character length of the input English text is \textbf{59.3}}. $\dagger$ indicates that the scores were directly cited from other research papers. }
    \label{multi30k_main}
\end{table*}

\paragraph{Comprehensive Performance Improvement from Speech-Text Fusion Input.}
Table \ref{multi30k_main} showcases the remarkable performance of our SMT-9B model, which expertly integrates both synthetic speech and text inputs. The results clearly demonstrate a substantial performance gain across all evaluated test sets. Specifically, for the \texttt{eng$\rightarrow$deu} task, our model attains impressive BLEU scores of 47.0, 41.8, and 40.3 on the Test2016, Test2017, and MSCOCO datasets, respectively. Similarly, for the \texttt{eng$\rightarrow$fra} task, it achieves high BLEU scores of 67.0, 62.1, and 55.3. These scores consistently and significantly outperform all text-only baselines. The clear advantage our approach holds provides compelling evidence that synthetic speech, as an auxiliary modality, can furnish crucial prosodic and contextual information that is not available in text alone, thereby effectively enhancing machine translation performance.

\paragraph{Competitive Advantage of Synthetic Speech in Multimodal Translation.}
The table clearly demonstrates the significant performance advantage of our proposed method, which leverages synthetic speech, over existing multimodal machine translation models that primarily rely on visual inputs. Our SMT-9B model establishes a new benchmark by achieving a state-of-the-art average BLEU score of 52.0. This score not only surpasses the performance of all previous methods but does so by a substantial margin, regardless of whether those models used authentic or synthetic images.
For a direct comparison, our model outperforms the best-performing image-based model by an impressive 2.1 points (which only achieved an average BLEU of 49.9). This result suggests that the speech modality is a rich and unique source of contextual information that is both distinct from and complementary to the visual modality. 

\paragraph{Comparative Analysis with Large-Scale Language Models.}
Although not shown in the table, our SMT-9B model, despite having a parameter count that is only 1/67th of the DeepSeek-V3-671B model, achieves superior translation performance. This result highlights the significant potential of multimodal learning: even a smaller model can achieve or surpass the performance of a much larger text-only model by effectively leveraging cross-modal information. This demonstrates that modality fusion can compensate for a lack of scale, offering a viable path for developing high-performance translation systems in resource-constrained environments.

\newpage

\label{sec:overall result}
\subsubsection{Experimental Results for Machine Translation}

\begin{table*}[t]
\small
\renewcommand{\arraystretch}{1.2}
\centering
\resizebox{1\textwidth}{!}{
\begin{tabular}{l| c c c c |cc}
\toprule
\multirow{2}{*}{\textbf{Models}} && \multicolumn{2}{c}{FLORES-200}&&  \multicolumn{2}{c}{WMT24++} \\
& {eng} $\rightarrow$ \texttt{27} &{jpn} $\rightarrow$ \texttt{27}&{kor} $\rightarrow$ \texttt{27}& {cmn} $\rightarrow$ \texttt{27}&{eng} $\rightarrow$ \texttt{22} &{eng} $\rightarrow$ \texttt{22} ($<$200)\\ \midrule
\rowcolor{gray!20}
\multicolumn{7}{c}{\textbf{Models based on Text}} \\ \midrule
DeepSeek-V3.1~\citep{guo2025deepseek} & \uline{39.3} / \uline{88.9} & \uline{26.1} / \uline{85.7} & \uline{27.7} / \uline{85.9} & \uline{27.5} / \uline{86.2} & 34.1 / \uline{83.6} & \uline{31.8} / \uline{83.4}\\
Gemma3-27B-it~\citep{team2025gemma} & 37.4 / 88.0 & 23.8 / 81.0& 25.0 / 81.2& 24.5 / 81.5 & \uline{34.3} / 82.9& \uline{31.8} / 82.6\\
NLLB-moe-54B~\citep{nllb2022} & 35.7 / 86.3& 21.8 / 81.7&23.6 / 83.7 & 22.8 / 82.1 & 25.4 / 76.9& 24.4 / 77.7\\
Qwen3-Next-80B-A3B~\citep{qwen3-next-blog-2025} & 34.5 / 86.6 &22.9 / 83.8& 23.9 / 83.9& 24.2 / 84.3 & 30.5 / 81.5 & 29.6 / 81.6\\
\midrule
\rowcolor{gray!20}
\multicolumn{7}{c}{\textbf{Models based on Text \& Synthetic Speech}} \\ \midrule
Baseline (Text only) & {39.7} / 88.3 &{26.6} / 85.4& 27.4 / 85.6& {27.5} / {85.7} & 33.9 / 82.7 & 32.1 / 82.9\\
SMT-9B &\colorbox{cvprblue!14}{40.4} / \colorbox{cvprblue!14}{89.5}&\colorbox{cvprblue!14}{27.3} / \colorbox{cvprblue!14}{86.9}&\colorbox{cvprblue!14}{28.3} / \colorbox{cvprblue!14}{87.1}& \colorbox{cvprblue!14}{28.3} / \colorbox{cvprblue!14}{87.4}& 33.4 / 83.0& \colorbox{cvprblue!14}{32.2} / \colorbox{cvprblue!14}{83.4} \\
\bottomrule
\end{tabular}
}
\raggedright{\scriptsize{\hspace*{0.5em}\underline{Underlined} denotes previous state-of-the-art models, while \colorbox{cvprblue!14}{highlighted} surpasses the previous models.}}
\caption{Translation Performance on FLORES-200 and WMT24++ (spBLEU / COMET) MT Benchmarks. {The average character length of the input English text is \textbf{130.4} for FLORES-200 and \textbf{191.3} for WMT24++. The notation $<200$ indicates that the input English text length is within 200 characters.} Detailed results are summarized in Tables~\ref{tab:wmt24}, and \ref{tab:flores200} in the Appendix.}
\label{tab:flores}
\end{table*}

\paragraph{Language Support.}
Our model exhibits strong language support, surpassing existing MMT models. Specifically, Table~\ref{tab:flores} details results for 108 translation directions on the \textbf{FLORES-200} benchmark, encompassing major source languages---{English (eng), Japanese (jpn), Korean (kor), and Chinese (cmn)---to 27 target languages. Furthermore, we evaluate on the \textbf{WMT24++} benchmark for \textbf{en$\rightarrow$22} directions.} The complete list of supported languages is provided in Table~\ref{tab:langs} in the Appendix.

\paragraph{Scalable Multilingualism.}
The consistent performance gain underscores our method's advantages: scalability and multilingual capability. As shown in the Table \ref{tab:flores}, our model not only performs exceptionally well on the \texttt{eng$\rightarrow$xx} task, but also delivers impressive gains on \texttt{jpn$\rightarrow$xx}, \texttt{kor$\rightarrow$xx}, and \texttt{cmn$\rightarrow$xx} directions. The average spBLEU scores for these language groups are 27.3, 28.3, and 28.3 respectively, all of which are the highest in their respective categories.

\begin{figure*}[b] \centering
\includegraphics[width=\textwidth]{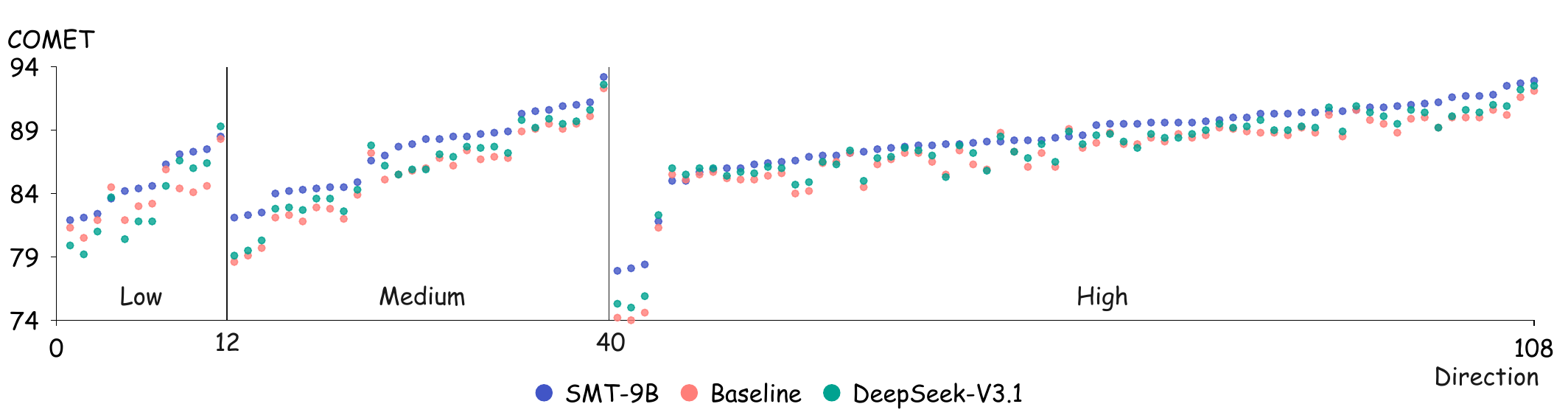} 
\caption{COMET Results by Resource Level, Categorized as Low, Medium, and High. Our model shows an improvement in translation scores, particularly for low-scoring translation directions.}
\label{resource}
\end{figure*}

\paragraph{SMT in Low-Scoring Directions.}
As shown in Figure~\ref{resource}, the \textbf{SMT-9B} model outperforms both the Baseline and DeepSeek models, particularly in low-resource translation directions like Khmer ($\text{khm}$), Lao ($\text{lao}$), and Burmese ($\text{mya}$), indicating its greater robustness in data-scarce language pairs. Beyond this, we note an underperforming high-resource language, Hindi (hin), whose translation metrics are lower than many low-resource counterparts.

{\paragraph{Translation Text Length.}
As shown in Table~\ref{tab:flores}, the WMT24++ dataset contains numerous extremely long texts, leading to noise (e.g., word omissions or duration exceeding $30s$) in the synthesized speech. Although the model's performance on the overall dataset is moderate, it exhibits good performance within the $<200$ range. More importantly, the model's performance does not significantly degrade compared to the baseline, even when receiving noisy speech input, which fully demonstrates the model's robustness.}

\newpage

\newpage

\subsubsection{Ablation Study}
\label{sec:ablation}  

\begin{table*}[t]
    \centering
    \setlength{\tabcolsep}{6pt} 
    \resizebox{\textwidth}{!}{
    \begin{tabular}{@{}c cccccccccc@{}}
    \toprule
    \multicolumn{3}{c}{\textbf{Input}} & \multicolumn{5}{c}{eng $\rightarrow$ \texttt{xx}} &\multicolumn{2}{r}{spBLEU / COMET $\uparrow$}\\ \cmidrule(lr){1-3} \cmidrule(lr){4-10}
    Text & AS& SS & ara & deu & fra &ind&jpn&tur& Avg.\\ \midrule
    \textcolor{cvprblue}{\ding{51}} &  &  & 37.7 / 86.3 & 45.2 / 88.0 & 32.1 / 86.9 & 47.9 / 91.5 & 31.5 / 90.7 & 36.7 / 88.8 & 38.5 / 88.7  \\
     & \textcolor{cvprblue}{\ding{51}} & &  32.6 / 82.2 & 36.6 / 82.2 & 27.9 / 82.6 & 36.8 / 85.9 &26.9 / 86.5& 29.3 / 83.6 & 31.7 / 83.8\\
     &  &\textcolor{cvprblue}{\ding{51}}& 34.1 / 83.5 & 39.0 / 84.0 & 28.9 / 83.8 & 36.9 / 87.4 & 27.1 / 87.4 & 30.3 / 85.0 & 32.7 / 85.4\\
         \textcolor{cvprblue}{\ding{51}} &\textcolor{cvprblue}{\ding{51}}&  & 40.1 / 86.8 &  46.5 / 88.3 &  33.6 / 87.4 &48.4 / 91.6&33.6 / 90.6&37.9 / 89.1&40.0 / 89.0\\
    \textcolor{cvprblue}{\ding{51}} && \textcolor{cvprblue}{\ding{51}} & 40.1 / 86.8 &  46.5 / 88.2 &  33.6 / 87.4 &48.5 / 91.6&33.5 / 90.7&37.8 / 89.1&40.0 / 89.0 \\
     \bottomrule
    \end{tabular}}
\caption{Ablation Study on the CoVoST-2 Benchmark. A comparison of configurations with different modality inputs. (AS denotes authentic speech; SS denotes synthetic speech)}

\label{ablation_loss}
\end{table*}
\begin{table*}[t]
    \centering
    \setlength{\tabcolsep}{6pt} 
    \resizebox{\textwidth}{!}{
    \begin{tabular}{@{}c cccccccccc@{}}
    \toprule
    \multirow{2}{*}{Models} & \multicolumn{5}{c}{eng $\rightarrow$ \texttt{xx}} &\multicolumn{2}{r}{spBLEU / COMET $\uparrow$}\\ \cmidrule(lr){2-8}
    & jpn & cmn & tha &khm&lao&mya& Avg.\\ \midrule
    Baseline & 33.3 / 91.3 & 41.6 / 89.2 & 42.5 / 88.7&24.1 / 84.2 & 31.5 / 84.7 & 20.1 / 88.1 &   32.2 / 87.7 \\
    SMT-9B & 35.2 / 92.7 &  42.6 / 91.2 &  44.1 / 90.3 &25.6 / 83.6&34.2 / 86.3&24.3 / 88.5& 34.3 / 88.8 \\
     w/o SE &  34.8 / 92.1& 42.3 / 89.3& 42.5 / 89.7& 23.0 / 81.7 & 31.7 / 84.3 & 23.4 / 86.8& 33.0 / 87.3\\
     \bottomrule
    \end{tabular}}
\caption{Ablation Study on Self-Evolution (SE) Mechanism on the FLORES-200 benchmark.}
\label{ablation_self}
\end{table*}

\paragraph{Authentic Speech vs. Synthetic Speech.}
As shown in Table \ref{ablation_loss}, experimental results reveal that the difference between authentic and synthetic speech has minimal impact on multimodal machine translation performance. Surprisingly, synthetic speech achieves better S2TT performance, likely due to the absence of background noise. Experimental results demonstrate strong semantic consistency between authentic and synthetic speech.

\paragraph{The Impact of the Self-Evolution Mechanism.} As shown in Table \ref{ablation_self}, we found that after MLLM pre-training, the model's performance on high-resource languages improved. However, due to the imbalance of multilingual data, the performance on low-resource languages like Khmer (khm), Lao (lao), and Burmese (mya) actually decreased on the COMET metric. Therefore, we introduced the self-evolution mechanism to enhance the model's performance on these low-resource directions.

\begin{figure*}[h] 
    \centering
    \includegraphics[scale=0.38]{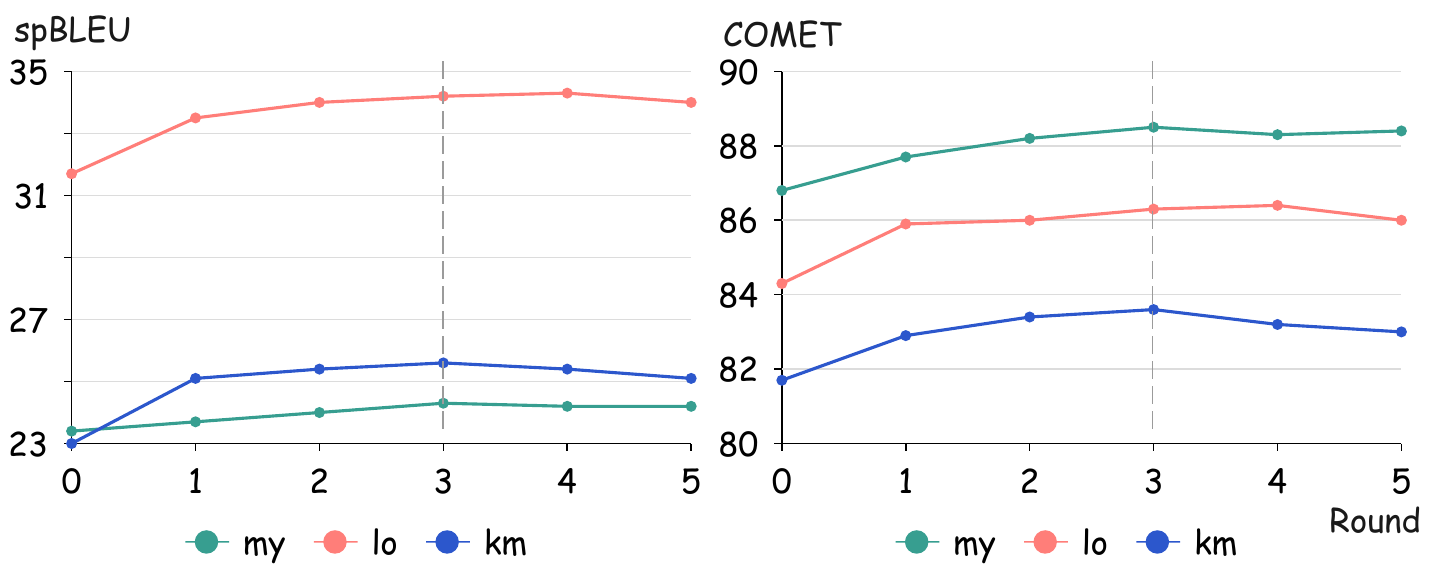} 
    \caption{Self-Evolution Rounds of spBLEU / COMET  (eng$\rightarrow$xx) on FLORES-200 benchmark.}
    \label{self}
\end{figure*}

\paragraph{Self-Evolution Rounds on Low-Resource Languages.}
Figure~\ref{self} shows the improvements from self-evolution for low-resource languages, with round~3 achieving best average gains of +1.9, +2.0, and +1.7~COMET on khm, lao, and mya, respectively. We observe that the first round yields the most significant improvement, later rounds give fewer benefits. The average improvement peaks at round 3 and then remains stable.

\begin{figure*}[t] \centering
\includegraphics[width=\textwidth]{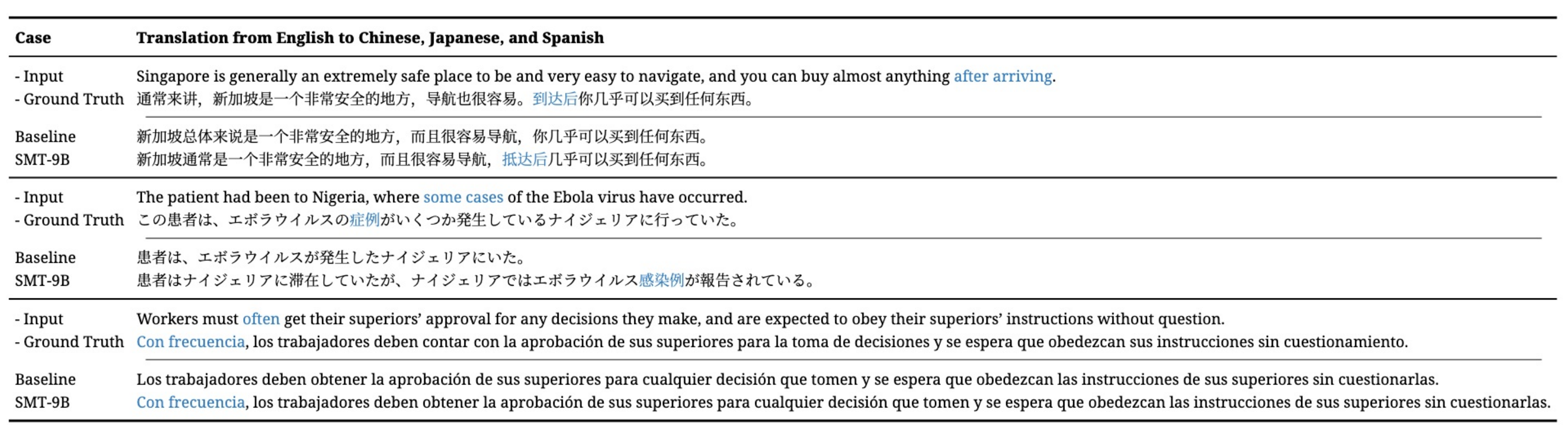} 
\caption{Case Study for Under-Translation. Having undergone speech pre-training, MLLMs align text words with speech. The SMT model, which receives this speech-text fusion input, is prevented from ignoring the input text, thereby mitigating omission errors.}
\label{case}
\end{figure*}

\paragraph{Human Evaluation for MT and SMT.}
Manual review of evaluation samples revealed that the performance gain from adding the speech modality is likely due to a reduction in \textbf{under-translation}, which decreased from 5.2\% to 3.5\%, as shown in Figure~\ref{case}. {The introduction of the speech modality provides prosodic cues as additional signals that effectively help correct the attention weighting, thereby mitigating this problem.}

\label{loss_ablation}

\newpage

\section{Related Work}

\paragraph{Multimodal Machine Translation.}
MMT research has primarily followed two distinct paths: image-based and image-free approaches. Image-based methods, exemplified by foundational work on the Multi30K dataset \citep{DBLP:conf/acl/ElliottFSS16}, utilize paired visual and textual data to improve translation quality. In contrast, image-free approaches emerged to tackle the challenges of data scarcity. These methods employ various techniques, such as target-end retrieval \citep{DBLP:conf/acl/HitschlerSR16}, multi-task learning \citep{DBLP:conf/ijcnlp/ElliottK17}, and even visual generation using advanced models like GANs and diffusion models \citep{rombach2022high}, to generate or retrieve supplementary information without relying on a pre-existing image dataset.

\paragraph{Multimodal Large Language Model.}
MLLMs \citep{chen2024slam,du-etal-2025-making,du2025mcatscalingmanytomanyspeechtotext} typically feature three core components: an LLM backbone, a modality encoder, and a modality adapter. Our framework specifically leverages this architecture to handle both speech and text. The speech encoder, inspired by models like Whisper \citep{radford2023robust}, is responsible for extracting rich speech features from the audio input. Following this, the speech adapter \citep{DBLP:conf/icml/0008LSH23} projects these features into the same hidden dimension as the LLM, enabling seamless integration. The processed speech features are then concatenated with the original text embeddings. This unified representation is fed into the LLM backbone, which processes both modalities jointly to generate the final translated text. 

\paragraph{Self-Evolution.}
The concept of self-evolution~\citep{liu2021survey} empowers models to autonomously acquire, refine, and learn from self-generated experiences. As outlined in recent surveys \citep{tao2024survey}, this process typically involves a four-phase iterative cycle: (1) experience acquisition, (2) experience refinement, (3) updating, and (4) evaluation. Each iteration is designed to achieve a specific evolutionary objective. In our implementation, the process begins with the experience acquisition phase, where we generate synthetic speech data. This is followed by a refinement phase that involves the annotation of positive and negative samples. This newly labeled data is then used to update the model, which is subsequently evaluated for its machine translation performance.

\section{Conclusion}
In this paper, we present the Speech-guided Machine Translation (SMT) framework, a novel approach that overcomes the limitations of traditional image-based multimodal translation.
Our framework integrates a TTS model with an MLLM, leveraging speech as a complementary modality to text. A key feature is the Self-Evolution Mechanism, which autonomously generates and refines training data. This significantly reduces the need for human-annotated data in low-resource languages, making the system more scalable and practical.
Our experiments show that SMT-9B achieves SOTA performance on benchmarks such as Multi30K and FLORES-200.

\section{Limitation}
\label{sec:limitation}
Unlike image-based methods, our speech-guided machine translation approach can cover a broader range of languages. However, we are still limited by the languages supported by the TTS models, as we need to synthesize speech from text. Although recent advancements in TTS technology have enabled the synthesis of dozens of languages, open-source TTS models still have limited language coverage.

\section{The Use of Large Language Models}
In this paper, LLMs are not used for ideation but are utilized for checking grammatical rules.

\section{Reproducibility statement}
All models and datasets tested in this research are open-source. 

\section*{Acknowledgements}
The research in this article is supported by the National Science and Technology Major Program (Grant No. 2024ZD01NL00101), the National Science Foundation of China (U22B2059, 62276083, 62506182), National Key Research and Development Program of China (2025YFE0200500), the Key Research and Development Program of Heilongjiang Province (2022ZX01A28) and the 5G Application Innovation Joint Research Institute’s Project (A003), and the Major Key Project of PCL.

\bibliography{iclr2026_conference}
\bibliographystyle{iclr2026_conference}

\newpage

\appendix
\section{Appendix}
\begin{table*}[h]
\centering
\small
\begin{tabular}{c | c | c | c | c | c}
\toprule
ISO-3 & Language & Script & Family & Subgrouping & Resource \\
\midrule
{ben} & Bemba &Latin &Atlantic-Congo& Benue-Congo & Low \\
{ces} & Czech & Latin &  Indo-European & Balto-Slavic & High \\
{cmn} & Chinese & Han & Sino-Tibetan & Sinitic & High \\ 
{deu} & German & Latin & Indo-European & Germanic & High \\
{eng} & English &  Latin & Indo-European &  Germanic & High \\
{fra} & French &  Latin & Indo-European &  Italic & High \\
{hau} & Hausa &Latin &Afro-Asiatic  &Chadic & Low \\
{hin} & Hindi & Devanagari &  Indo-European & Indo-Aryan & High \\
{lav} & Latvian & Latin& Indo-European &Balto-Slavic & High \\
{spa} & Spanish & Latin & Indo-European & Italic & High \\
{tur} & Turkish & Latin & Turkic &  Common Turkic & High \\ \bottomrule
\hline
\end{tabular}
\caption{11 Languages Supported by Image-Guided MMT datasets. The resource of each language is determined according to the taxonomy classes by \citep{joshi2020state}.}
\label{tab:mmt_langs}
\end{table*}

\begin{table*}[h]
\centering
\small
\begin{tabular}{c | c | c | c | c | c}
\toprule
ISO-3 & Language & Script & Family & Subgrouping & Resource \\
\midrule
{ara} & Arabic & Arabic & Afro-Asiatic & Semitic & High \\
{ben} & Bengali & Bengali & Indo-European & Indo-Aryan & Med \\
{ces} & Czech & Latin &  Indo-European & Balto-Slavic & High \\
{cmn} & Chinese & Han & Sino-Tibetan & Sinitic & High \\ 
{deu} & German & Latin & Indo-European & Germanic & High \\
{eng} & English &  Latin & Indo-European &  Germanic & High \\
{fas} & Persian & Arabic & Indo-European & Iranian & High \\
{fra} & French &  Latin & Indo-European &  Italic & High \\
{heb} & Hebrew & Hebrew & Afro-Asiatic & Semitic & Med \\
{hin} & Hindi & Devanagari &  Indo-European & Indo-Aryan & High \\
{ind} & Indonesian &  Latin &  Austronesian & Malayo-Polynesian & Med \\
{ita} & Italian & Latin & Indo-European &  Italic & High \\
{jpn} & Japanese & Japanese &  Japonic &  Japanesic & High \\
{khm} & Khmer & Khmer & Austroasiatic & Khmeric & Low \\
{kor} & Korean & Hangul &  Koreanic &  Korean & High \\
{lao} & Lao & Lao & Tai-Kadai & Kam-Tai & Low \\
{msa} & Malay &  Latin & Austronesian & Malayo-Polynesian  & Med \\
{mya} & Burmese & Myanmar & Sino-Tibetan & Burmo-Qiangic & Low \\
{nld} & Dutch & Latin &  Indo-European &  Germanic & High \\
{pol} & Polish &  Latin & Indo-European & Balto-Slavic & High \\
{por} & Portuguese &  Latin & Indo-European & Italic & High \\
{rus} & Russian & Cyrillic & Indo-European &  Balto-Slavic & High \\
{spa} & Spanish & Latin & Indo-European & Italic & High \\
{tgl} & Tagalog & Latin &  Austronesian & Malayo-Polynesian & Med \\
{tha} & Thai & Thai &  Tai-Kadai &  Kam-Tai & Med \\
{tur} & Turkish & Latin & Turkic &  Common Turkic & High \\
{urd} & Urdu &  Arabic & Indo-European & Indo-Aryan & Med \\
{vie} & Vietnamese & Latin & Austroasiatic & Vietic & High \\ \bottomrule
\hline
\end{tabular}
\caption{28 Languages Supported by Our Model.
 The resource of each language is determined according to the taxonomy classes by \citep{joshi2020state}.}
\label{tab:langs}
\end{table*}

\begin{table*}[h]
\centering
\tiny
\resizebox{\textwidth}{!}{%
\begin{tabular}{lcccccc}
\toprule
\textbf{Model} & \textbf{Task} & \textbf{Description} & \textbf{Dataset} & \textbf{Split}& \textbf{Data Size} & \textbf{Metric} \\
\midrule
\multirow{4}{*}{MLLM} & \multirow{2}{*}{ASR} & \multirow{2}{*}{Automatic Speech Recognition} & FLEURS$^{\dagger}$ & train &$\sim$160h& \multirow{2}{*}{WER $\downarrow$} \\
& & & Common Voice 19 & train &$\sim$3000h& \\
\cmidrule(lr){2-7}
& \multirow{2}{*}{SMT} & \multirow{1}{*}{Speech-Guided} & FLEURS$^{\dagger}$ & train &$\sim$160h & spBLEU / COMET $\uparrow$ \\
& &{ Multimodal Machine Translation} & Multi30K & train &$\sim$40h& BLEU / COMET $\uparrow$ \\
\bottomrule
\end{tabular}
}
\caption{Summary of Training Datasets for SMT Models. {Data size refers to the actual amount used for training, as we removed some overly long samples. $\dagger$ indicates that we performed data cleaning on the dataset. Since there is an overlap between the FLEURS and FLORES datasets, we removed the overlapping portions from the FLEURS training set.} }
\label{tab:train}
\end{table*}

\begin{table*}[h]
\centering
\tiny
\resizebox{\textwidth}{!}{%
\begin{tabular}{lcccc}
\toprule
\textbf{Task}                  & \textbf{Description}                   & \textbf{Dataset}                    & \textbf{Split} & \textbf{Metric} \\
\midrule
\multirow{2}{*}{MT} & \multirow{2}{*}{Machine Translation} & FLORES-200  & devtest     & \multirow{2}{*}{spBLEU / COMET $\uparrow$}    \\
&  & WMT24++  & test     &     \\
                     \multirow{2}{*}{MMT}                  & \multirow{2}{*}{Multimodal Machine Translation}                   & \multirow{2}{*}{Multi30K}                   & \multirow{2}{*}{test}           & \multirow{2}{*}{BLEU / COMET $\uparrow$}                \\
 \\

\multirow{2}{*}{S2TT}                 & \multirow{2}{*}{Speech-to-Text Translation}                   & \multirow{2}{*}{CoVoST-2}             & \multirow{2}{*}{test}           & \multirow{2}{*}{spBLEU / COMET $\uparrow$} \\
\\
\bottomrule         
\end{tabular}
}
\caption{Summary of Evaluation Benchmarks.}
\label{tab:benchmark}
\end{table*}

\begingroup

\begin{table}[h]\tiny
    \centering
            \setlength{\tabcolsep}{10pt} 
    \renewcommand{\arraystretch}{1}
        \begin{tabular}{l|cccccc}
        \toprule
        \multirow{2}{*}{Direction} 	&DeepSeek 		&	Gemma3			&		NLLB-moe	&	  Qwen3-Next  				&	     \multirow{2}{*}{Baseline}		&		SMT			\\ 
           & -v3.1&-27B&-54B& -80B-A3B & &-9B\\ \midrule
eng $\rightarrow$ ara	&	20.0 / 78.5 & 	20.0 / 78.1 & 	18.5 / 74.6 & 	19.4 / 77.6 & 	19.4 / 77.3 & 	19.1 / 77.5	\\
eng $\rightarrow$ ben	&	25.9 / 83.3 & 	25.4 / 82.7 & 	23.5 / 79.7 & 	16.2 / 78.5 & 	24.8 / 82.1 & 	23.6 / 80.7	\\
eng $\rightarrow$ ces	&	36.3 / 85.9 & 	36.3 / 84.6 & 	23.4 / 79.0 & 	30.3 / 82.2 & 	36.4 / 85.1 & 	35.8 / 85.5	\\
eng $\rightarrow$ cmn 	&	34.3 / 84.9 & 	36.4 / 83.5 & 	18.0 / 69.5 & 	37.4 / 84.9 & 	36.9 / 83.8 & 	36.5 / 85.3	\\
eng $\rightarrow$ deu	&	37.9 / 82.6 & 	37.9 / 81.9 & 	28.5 / 76.3 & 	36.2 / 81.7 & 	37.7 / 82.3 & 	37.3 / 82.3	\\
eng $\rightarrow$ fas	&	29.9 / 83.1 & 	32.9 / 83.1 & 	25.8 / 78.0 & 	27.7 / 80.7 & 	32.1 / 83.1 & 	31.9 / 83.8	\\
eng $\rightarrow$ fra	&	48.1 / 82.7 & 	47.4 / 82.2 & 	34.8 / 75.5 & 	45.0 / 82.0 & 	44.3 / 82.2 & 	45.0 / 81.9	\\
eng $\rightarrow$ heb	&	37.4 / 82.6 & 	36.6 / 82.3 & 	33.9 / 79.4 & 	26.8 / 76.7 & 	38.8 / 83.5 & 	38.3 / 84.5	\\
eng $\rightarrow$ hin	&	19.6 / 74.0 & 	19.5 / 73.4 & 	16.0 / 65.5 & 	12.6 / 68.5 & 	19.3 / 71.0 & 	19.6 / 70.2	\\
eng $\rightarrow$ ind	&	38.2 / 86.8 & 	37.6 / 86.3 & 	30.6 / 80.8 & 	36.6 / 86.0 & 	37.3 / 85.3 & 	37.2 / 86.0	\\
eng $\rightarrow$ ita	&	45.2 / 84.7 & 	46.2 / 84.4 & 	33.3 / 78.6 & 	41.9 / 83.7 & 	45.0 / 84.6 & 	44.2 / 85.2	\\
eng $\rightarrow$ jpn	&	25.4 / 87.6 & 	24.0 / 86.4 & 	11.7 / 79.1 & 	22.6 / 86.9 & 	22.5 / 85.7 & 	22.5 / 85.9	\\
eng $\rightarrow$ kor	&	27.1 / 87.3 & 	26.9 / 86.4 & 	20.7 / 81.9 & 	23.9 / 86.1 & 	26.0 / 85.6 & 	25.0 / 85.4	\\
eng $\rightarrow$ nld	&	40.4 / 84.4 & 	39.3 / 83.7 & 	28.5 / 77.8 & 	35.8 / 82.7 & 	38.7 / 84.6 & 	37.5 / 84.3	\\
eng $\rightarrow$ pol	&	30.5 / 84.8 & 	29.2 / 83.9 & 	18.0 / 77.3 & 	25.6 / 81.7 & 	29.4 / 83.8 & 	28.7 / 84.8	\\
eng $\rightarrow$ por	&	40.7 / 83.4 & 	40.0 / 82.9 & 	28.7 / 77.2 & 	38.6 / 82.7 & 	39.5 / 83.0 & 	39.3 / 83.4	\\
eng $\rightarrow$ rus	&	29.6 / 83.4 & 	31.4 / 82.7 & 	23.2 / 76.6 & 	28.8 / 81.9 & 	29.2 / 81.9 & 	29.9 / 83.5	\\
eng $\rightarrow$ spa	&	48.4 / 83.7 & 	48.7 / 83.6 & 	36.0 / 77.7 & 	46.2 / 83.0 & 	48.5 / 83.7 & 	46.1 / 83.8	\\
eng $\rightarrow$ tha	&	32.6 / 85.1 & 	33.8 / 84.8 & 	22.3 / 77.9 & 	29.6 / 83.5 & 	32.4 / 82.8 & 	31.7 / 83.7	\\
eng $\rightarrow$ tra	&	36.0 / 85.5 & 	36.6 / 84.3 & 	27.0 / 79.0 & 	30.6 / 83.0 & 	36.6 / 84.6 & 	36.3 / 84.2	\\
eng $\rightarrow$ urd	&	30.5 / 79.8 & 	30.3 / 79.0 & 	29.0 / 73.7 & 	23.5 / 75.8 & 	33.3 / 79.8 & 	32.4 / 80.5	\\
eng $\rightarrow$ vie	&	36.6 / 84.8 & 	37.2 / 84.1 & 	26.5 / 77.7 & 	35.9 / 83.9 & 	37.6 / 83.7 & 	37.1 / 84.2	\\ \midrule
Avg.	&	34.1 / 83.6 & 	34.3 / 82.9 & 	25.4 / 76.9 & 	30.5 / 81.5 & 	33.9 / 82.7 & 	33.4 / 83.0	\\
        \bottomrule
        \end{tabular}
    
    \caption{spBLEU / COMET Scores on the WMT24++ Benchmark.}
    \label{tab:wmt24}
    \end{table}

\begin{table}[h]\tiny
    \centering
    \tiny
            \setlength{\tabcolsep}{10pt} 
    \renewcommand{\arraystretch}{0.6}
        \begin{tabular}{l|cccccc}
        \toprule
        \multirow{2}{*}{Direction} 	&DeepSeek 		&	Gemma3			&		NLLB-moe	&	  Qwen3-Next  				&	     \multirow{2}{*}{Baseline}		&		SMT			\\ 
           & -v3.1&-27B&-54B& -80B-A3B & &-9B\\ \midrule

eng $\rightarrow$ ara	&	41.6 / 88.1 & 	41.7 / 87.8 & 	41.8 / 86.8 & 	38.3 / 87.1 & 	42.9 / 87.8 & 	42.6 / 89.5\\	
eng $\rightarrow$ ben	&	33.6 / 87.8 & 	30.0 / 86.6 & 	34.5 / 86.4 & 	28.0 / 85.6 & 	34.8 / 86.6 & 	34.3 / 86.6\\	
eng $\rightarrow$ ces	&	44.0 / 92.5 & 	41.4 / 91.3 & 	40.3 / 90.9 & 	37.9 / 90.2 & 	42.7 / 91.6 & 	43.1 / 92.9\\	
eng $\rightarrow$ cmn	&	35.7 / 89.2 & 	37.2 / 88.8 & 	22.4 / 78.0 & 	37.0 / 89.2 & 	41.6 / 89.2 & 	42.6 / 91.2\\	
eng $\rightarrow$ deu	&	48.5 / 89.0 & 	46.9 / 88.7 & 	43.8 / 87.1 & 	46.2 / 88.5 & 	47.1 / 88.5 & 	47.8 / 89.7\\	
eng $\rightarrow$ fas	&	35.1 / 89.0 & 	35.3 / 88.7 & 	34.4 / 87.2 & 	30.9 / 86.8 & 	38.7 / 88.9 & 	38.3 / 90.3\\	
eng $\rightarrow$ fra	&	56.3 / 89.2 & 	55.6 / 88.8 & 	54.6 / 87.7 & 	55.1 / 88.8 & 	57.7 / 89.1 & 	57.1 / 90.0\\	
eng $\rightarrow$ heb	&	47.8 / 89.7 & 	45.4 / 89.1 & 	45.0 / 88.4 & 	33.1 / 83.4 & 	46.3 / 89.3 & 	46.8 / 91.0\\	
eng $\rightarrow$ hin	&	37.9 / 82.3 & 	36.8 / 81.7 & 	38.6 / 80.7 & 	31.9 / 79.9 & 	41.3 / 81.1 & 	41.0 / 81.8\\	
eng $\rightarrow$ ind	&	50.0 / 92.6 & 	49.5 / 92.0 & 	48.1 / 91.1 & 	48.7 / 92.1 & 	52.6 / 92.2 & 	52.4 / 93.2\\	
eng $\rightarrow$ ita	&	39.1 / 89.3 & 	39.1 / 89.4 & 	37.1 / 88.1 & 	37.7 / 89.0 & 	38.8 / 89.3 & 	39.4 / 90.4\\	
eng $\rightarrow$ jpn	&	33.9 / 92.2 & 	32.6 / 91.8 & 	18.8 / 88.1 & 	29.0 / 91.7 & 	33.3 / 91.3 & 	35.2 / 92.7\\	
eng $\rightarrow$ khm	&	23.8 / 83.7 & 	17.7 / 81.3 & 	22.0 / 79.5 & 	15.0 / 76.3 & 	24.1 / 84.2 & 	25.6 / 83.6\\	
eng $\rightarrow$ kor	&	29.5 / 90.8 & 	28.8 / 90.3 & 	25.4 / 89.0 & 	26.2 / 90.0 & 	30.4 / 90.1 & 	30.1 / 90.5\\	
eng $\rightarrow$ lao	&	30.0 / 84.6 & 	27.7 / 83.1 & 	29.1 / 83.4 & 	17.0 / 73.8 & 	31.5 / 84.7 & 	34.2 / 86.3\\	
eng $\rightarrow$ msa	&	45.2 / 90.6 & 	37.6 / 86.8 & 	44.4 / 88.7 & 	39.7 / 89.7 & 	47.0 / 90.9 & 	47.4 / 91.2\\	
eng $\rightarrow$ mya	&	24.0 / 89.3 & 	15.2 / 85.7 & 	16.1 / 83.7 & 	14.7 / 82.2 & 	20.1 / 88.2 & 	24.3 / 88.5\\	
eng $\rightarrow$ nld	&	36.6 / 88.7 & 	35.4 / 88.5 & 	34.6 / 87.3 & 	33.9 / 87.9 & 	37.5 / 88.8 & 	37.2 / 89.5\\	
eng $\rightarrow$ pol	&	35.3 / 90.6 & 	34.0 / 90.2 & 	30.9 / 88.6 & 	30.8 / 88.7 & 	33.5 / 89.9 & 	34.1 / 91.7\\	
eng $\rightarrow$ por	&	55.3 / 90.4 & 	55.3 / 90.4 & 	51.0 / 88.8 & 	53.9 / 90.1 & 	53.2 / 90.0 & 	55.4 / 91.1\\	
eng $\rightarrow$ rus	&	43.0 / 90.9 & 	41.2 / 90.1 & 	38.8 / 88.8 & 	40.2 / 90.1 & 	41.4 / 90.1 & 	41.6 / 92.5\\	
eng $\rightarrow$ spa	&	34.4 / 87.3 & 	33.9 / 87.2 & 	32.3 / 85.9 & 	33.4 / 87.0 & 	35.5 / 87.2 & 	36.5 / 88.2\\	
eng $\rightarrow$ tgl	&	39.5 / 86.2 & 	38.9 / 85.9 & 	37.4 / 84.5 & 	30.1 / 82.5 & 	38.2 / 84.5 & 	41.0 / 87.0\\	
eng $\rightarrow$ tha	&	44.8 / 89.8 & 	42.8 / 89.4 & 	32.1 / 83.7 & 	40.8 / 88.8 & 	42.5 / 88.7 & 	44.1 / 90.3\\	
eng $\rightarrow$ tur	&	42.1 / 90.9 & 	40.2 / 90.5 & 	39.5 / 89.2 & 	35.5 / 89.3 & 	42.2 / 90.6 & 	41.7 / 90.6\\	
eng $\rightarrow$ urd	&	30.3 / 84.3 & 	27.6 / 83.0 & 	28.8 / 81.0 & 	23.8 / 80.6 & 	30.9 / 83.9 & 	30.7 / 84.9\\	
eng $\rightarrow$ vie	&	44.4 / 90.4 & 	43.6 / 90.0 & 	42.5 / 87.9 & 	42.8 / 89.8 & 	46.7 / 90.0 & 	46.4 / 91.7\\  \midrule	
jpn $\rightarrow$ ara	&	27.3 / 84.7 & 	26.8 / 84.1 & 	23.3 / 80.8 & 	24.1 / 83.6 & 	26.5 / 84.0 & 	26.8 / 86.6\\	
jpn $\rightarrow$ ben	&	23.7 / 82.7 & 	0.6 / 49.0 & 	20.9 / 79.4 & 	19.7 / 80.4 & 	24.1 / 81.8 & 	24.0 / 84.3\\	
jpn $\rightarrow$ ces	&	26.7 / 90.4 & 	25.9 / 89.1 & 	20.2 / 86.1 & 	23.2 / 89.1 & 	25.8 / 89.8 & 	27.2 / 90.8\\	
jpn $\rightarrow$ cmn	&	27.5 / 88.4 & 	27.9 / 88.2 & 	15.1 / 74.8 & 	27.2 / 88.3 & 	32.0 / 88.7 & 	33.3 / 89.6\\	
jpn $\rightarrow$ deu	&	29.2 / 85.7 & 	28.7 / 85.3 & 	23.9 / 81.9 & 	27.1 / 85.0 & 	28.2 / 85.1 & 	28.8 / 86.0\\	
jpn $\rightarrow$ eng	&	32.7 / 88.5 & 	33.4 / 88.5 & 	33.2 / 87.4 & 	32.4 / 88.5 & 	36.9 / 88.8 & 	37.8 / 88.1\\	
jpn $\rightarrow$ fas	&	22.7 / 85.3 & 	15.3 / 67.6 & 	18.5 / 79.9 & 	20.4 / 83.9 & 	24.4 / 85.5 & 	25.4 / 87.9\\	
jpn $\rightarrow$ fra	&	34.0 / 86.0 & 	34.0 / 85.8 & 	29.8 / 83.0 & 	32.2 / 85.4 & 	32.8 / 85.7 & 	34.7 / 85.9\\	
jpn $\rightarrow$ heb	&	27.4 / 85.5 & 	27.0 / 85.4 & 	21.0 / 79.2 & 	19.9 / 80.4 & 	27.1 / 85.5 & 	27.9 / 87.7\\	
jpn $\rightarrow$ hin	&	24.2 / 75.3 & 	24.0 / 74.8 & 	21.0 / 71.4 & 	19.8 / 73.2 & 	24.1 / 74.2 & 	24.7 / 77.9\\	
jpn $\rightarrow$ ind	&	28.1 / 89.2 & 	28.4 / 88.4 & 	25.3 / 87.0 & 	27.1 / 88.8 & 	30.4 / 89.1 & 	30.9 / 90.5\\	
jpn $\rightarrow$ ita	&	27.3 / 87.4 & 	26.9 / 87.2 & 	22.1 / 84.0 & 	25.1 / 86.9 & 	26.3 / 87.2 & 	26.7 / 87.8\\	
jpn $\rightarrow$ khm	&	18.5 / 79.9 & 	13.7 / 77.1 & 	16.4 / 77.7 & 	12.0 / 73.8 & 	21.1 / 81.3 & 	19.1 / 81.9\\	
jpn $\rightarrow$ kor	&	23.6 / 88.7 & 	23.5 / 88.2 & 	19.8 / 84.8 & 	21.6 / 88.5 & 	23.5 / 88.4 & 	23.5 / 89.6\\	
jpn $\rightarrow$ lao	&	19.8 / 80.4 & 	20.0 / 79.6 & 	21.9 / 80.2 & 	12.0 / 70.6 & 	24.7 / 81.9 & 	24.2 / 84.2\\	
jpn $\rightarrow$ msa	&	25.9 / 87.1 & 	21.9 / 84.0 & 	23.2 / 84.7 & 	21.4 / 86.4 & 	27.3 / 86.8 & 	28.6 / 88.3\\	
jpn $\rightarrow$ mya	&	17.3 / 86.0 & 	0.3 / 27.8 & 	12.1 / 81.2 & 	11.7 / 78.9 & 	16.5 / 84.1 & 	19.3 / 87.3\\	
jpn $\rightarrow$ nld	&	24.7 / 86.3 & 	24.9 / 86.3 & 	20.2 / 82.1 & 	23.2 / 85.8 & 	25.8 / 86.5 & 	25.9 / 87.0\\	
jpn $\rightarrow$ pol	&	24.6 / 89.5 & 	25.1 / 89.3 & 	18.9 / 84.7 & 	21.8 / 88.2 & 	23.9 / 89.2 & 	24.2 / 89.8\\	
jpn $\rightarrow$ por	&	30.4 / 87.4 & 	31.4 / 87.3 & 	26.9 / 84.6 & 	29.4 / 87.1 & 	31.3 / 87.2 & 	32.3 / 87.2\\	
jpn $\rightarrow$ rus	&	28.1 / 89.0 & 	27.1 / 87.9 & 	23.8 / 86.0 & 	26.2 / 88.4 & 	26.9 / 88.6 & 	27.4 / 90.3\\	
jpn $\rightarrow$ spa	&	24.8 / 86.0 & 	23.8 / 85.5 & 	20.9 / 83.5 & 	22.9 / 85.5 & 	24.3 / 85.5 & 	25.3 / 85.0\\	
jpn $\rightarrow$ tgl	&	24.1 / 82.8 & 	23.4 / 82.0 & 	18.8 / 78.7 & 	17.8 / 80.0 & 	23.1 / 82.1 & 	24.9 / 84.0\\	
jpn $\rightarrow$ tha	&	35.5 / 86.9 & 	35.1 / 86.8 & 	24.0 / 79.9 & 	32.8 / 86.3 & 	33.9 / 86.2 & 	34.8 / 88.5\\	
jpn $\rightarrow$ tur	&	25.9 / 87.0 & 	25.6 / 86.7 & 	21.4 / 82.2 & 	21.9 / 85.5 & 	26.1 / 86.5 & 	26.4 / 87.8\\	
jpn $\rightarrow$ urd	&	20.6 / 79.1 & 	18.7 / 78.1 & 	19.3 / 75.9 & 	15.6 / 76.1 & 	20.1 / 78.6 & 	20.8 / 82.1\\	
jpn $\rightarrow$ vie	&	29.7 / 88.4 & 	30.4 / 88.3 & 	26.7 / 85.5 & 	28.8 / 88.0 & 	31.4 / 88.1 & 	32.2 / 89.6\\ 	
kor $\rightarrow$ ara	&	28.4 / 84.9 & 	28.5 / 85.0 & 	25.9 / 83.6 & 	25.3 / 84.1 & 	26.9 / 84.2 & 	27.4 / 86.9\\	 \midrule
kor $\rightarrow$ ben	&	25.2 / 82.9 & 	0.8 / 30.2 & 	22.7 / 81.2 & 	20.7 / 80.6 & 	24.9 / 82.3 & 	24.7 / 84.2\\	
kor $\rightarrow$ ces	&	29.0 / 90.1 & 	27.1 / 89.3 & 	23.4 / 87.9 & 	23.5 / 88.5 & 	27.2 / 89.5 & 	27.8 / 90.8\\	
kor $\rightarrow$ cmn	&	28.4 / 87.6 & 	29.3 / 87.4 & 	19.1 / 80.4 & 	28.5 / 87.6 & 	32.4 / 87.9 & 	33.8 / 89.5\\	
kor $\rightarrow$ deu	&	30.6 / 85.6 & 	29.3 / 84.7 & 	25.5 / 83.2 & 	27.7 / 84.9 & 	29.1 / 85.1 & 	29.8 / 86.3\\	
kor $\rightarrow$ eng	&	35.4 / 88.9 & 	35.9 / 88.9 & 	34.3 / 87.9 & 	34.9 / 88.8 & 	39.1 / 89.1 & 	40.3 / 88.5\\	
kor $\rightarrow$ fas	&	24.5 / 85.8 & 	24.0 / 84.0 & 	22.3 / 84.2 & 	21.4 / 84.1 & 	25.6 / 85.9 & 	26.2 / 88.1\\	
kor $\rightarrow$ fra	&	35.3 / 85.4 & 	35.4 / 85.2 & 	31.9 / 83.6 & 	33.7 / 84.9 & 	33.6 / 85.2 & 	35.9 / 86.0\\	
kor $\rightarrow$ heb	&	28.4 / 85.9 & 	29.1 / 85.8 & 	25.6 / 84.4 & 	20.2 / 80.7 & 	28.1 / 85.8 & 	28.8 / 87.9\\	
kor $\rightarrow$ hin	&	25.2 / 75.0 & 	25.3 / 74.8 & 	23.5 / 72.5 & 	21.1 / 73.2 & 	24.9 / 74.0 & 	25.7 / 78.1\\	
kor $\rightarrow$ ind	&	30.7 / 89.9 & 	31.0 / 89.7 & 	27.4 / 88.5 & 	28.9 / 89.4 & 	31.7 / 89.5 & 	32.0 / 90.6\\	
kor $\rightarrow$ ita	&	27.7 / 86.9 & 	24.2 / 82.3 & 	23.4 / 84.9 & 	25.5 / 86.4 & 	26.8 / 86.7 & 	26.9 / 87.6\\	
kor $\rightarrow$ jpn	&	26.7 / 90.1 & 	27.8 / 90.4 & 	15.6 / 86.8 & 	25.1 / 90.1 & 	26.9 / 90.0 & 	28.8 / 91.6\\	
kor $\rightarrow$ khm	&	19.3 / 79.2 & 	12.7 / 75.1 & 	17.5 / 79.1 & 	12.1 / 72.9 & 	21.3 / 80.5 & 	20.0 / 82.1\\	
kor $\rightarrow$ lao	&	22.7 / 81.8 & 	21.0 / 80.3 & 	22.1 / 81.9 & 	12.5 / 72.3 & 	25.7 / 83.0 & 	24.9 / 84.4\\	
kor $\rightarrow$ msa	&	27.4 / 87.7 & 	22.6 / 84.6 & 	24.2 / 86.0 & 	22.7 / 86.8 & 	28.5 / 87.4 & 	29.4 / 88.5\\	
kor $\rightarrow$ mya	&	19.5 / 86.6 & 	1.2 / 38.1 & 	12.1 / 82.9 & 	12.1 / 80.0 & 	15.8 / 84.4 & 	19.2 / 87.1\\	
kor $\rightarrow$ nld	&	26.0 / 86.5 & 	25.7 / 85.7 & 	22.9 / 84.6 & 	23.8 / 85.5 & 	26.0 / 86.4 & 	26.3 / 87.0\\	
kor $\rightarrow$ pol	&	25.8 / 89.3 & 	25.5 / 88.9 & 	21.5 / 86.9 & 	22.6 / 87.8 & 	24.4 / 88.9 & 	24.6 / 90.0\\	
kor $\rightarrow$ por	&	32.7 / 87.6 & 	28.4 / 85.5 & 	27.9 / 85.4 & 	31.3 / 87.2 & 	32.2 / 87.2 & 	33.6 / 87.7\\	
kor $\rightarrow$ rus	&	29.5 / 88.9 & 	28.3 / 88.1 & 	25.5 / 87.3 & 	26.7 / 88.3 & 	27.5 / 88.5 & 	28.2 / 90.5\\	
kor $\rightarrow$ spa	&	24.8 / 85.5 & 	24.1 / 85.2 & 	21.3 / 83.6 & 	23.0 / 85.0 & 	24.6 / 85.1 & 	25.6 / 85.0\\	
kor $\rightarrow$ tgl	&	25.1 / 83.6 & 	24.9 / 83.4 & 	21.2 / 81.3 & 	18.8 / 80.8 & 	23.8 / 82.9 & 	25.8 / 84.4\\	
kor $\rightarrow$ tha	&	36.6 / 87.6 & 	35.5 / 87.5 & 	26.2 / 83.3 & 	33.4 / 86.8 & 	33.5 / 86.7 & 	35.0 / 88.7\\	
kor $\rightarrow$ tur	&	28.4 / 87.2 & 	27.4 / 86.8 & 	24.0 / 84.8 & 	23.6 / 85.4 & 	27.2 / 86.3 & 	27.9 / 88.0\\	
kor $\rightarrow$ urd	&	21.4 / 79.5 & 	19.6 / 78.6 & 	20.3 / 77.2 & 	16.3 / 76.2 & 	21.0 / 79.1 & 	21.4 / 82.3\\	
kor $\rightarrow$ vie	&	32.4 / 88.7 & 	31.8 / 88.3 & 	29.1 / 87.0 & 	29.8 / 88.0 & 	32.3 / 88.4 & 	33.1 / 89.6\\ 	 \midrule
cmn $\rightarrow$ ara	&	28.4 / 85.0 & 	28.0 / 84.8 & 	25.4 / 82.5 & 	26.0 / 84.2 & 	27.6 / 84.5 & 	28.0 / 87.3\\	
cmn $\rightarrow$ ben	&	24.7 / 83.6 & 	10.2 / 62.2 & 	21.2 / 79.9 & 	20.6 / 81.8 & 	24.4 / 82.8 & 	24.3 / 84.5\\	
cmn $\rightarrow$ ces	&	29.1 / 90.6 & 	27.7 / 89.5 & 	21.7 / 85.2 & 	25.7 / 89.1 & 	28.0 / 89.9 & 	28.9 / 91.0\\	
cmn $\rightarrow$ deu	&	30.7 / 86.1 & 	29.2 / 85.5 & 	24.2 / 81.6 & 	28.6 / 85.6 & 	29.5 / 85.4 & 	30.0 / 86.4\\	
cmn $\rightarrow$ eng	&	35.0 / 87.9 & 	35.0 / 87.6 & 	35.0 / 86.6 & 	35.5 / 87.8 & 	36.1 / 87.6 & 	39.2 / 88.6\\	
cmn $\rightarrow$ fas	&	24.9 / 86.5 & 	17.3 / 72.4 & 	20.5 / 81.7 & 	22.2 / 84.8 & 	26.4 / 86.1 & 	27.0 / 88.4\\	
cmn $\rightarrow$ fra	&	36.1 / 86.0 & 	35.7 / 85.6 & 	31.4 / 82.2 & 	34.7 / 85.6 & 	34.6 / 85.6 & 	36.9 / 86.5\\	
cmn $\rightarrow$ heb	&	27.9 / 85.9 & 	27.3 / 85.5 & 	24.1 / 82.6 & 	21.0 / 80.9 & 	28.8 / 86.0 & 	29.4 / 88.3\\	
cmn $\rightarrow$ hin	&	25.2 / 75.9 & 	24.8 / 75.4 & 	21.9 / 71.6 & 	21.4 / 73.9 & 	25.3 / 74.6 & 	26.4 / 78.4\\	
cmn $\rightarrow$ ind	&	30.5 / 89.5 & 	30.0 / 88.7 & 	27.0 / 87.0 & 	29.5 / 89.2 & 	32.2 / 89.1 & 	32.3 / 90.9\\	
cmn $\rightarrow$ ita	&	29.1 / 87.9 & 	28.7 / 87.6 & 	23.2 / 83.6 & 	26.9 / 87.3 & 	27.1 / 87.2 & 	28.2 / 88.2\\	
cmn $\rightarrow$ jpn	&	26.2 / 91.0 & 	25.8 / 90.7 & 	14.5 / 84.7 & 	23.3 / 90.7 & 	25.3 / 90.6 & 	27.3 / 91.8\\	
cmn $\rightarrow$ khm	&	19.7 / 81.0 & 	13.9 / 77.3 & 	16.1 / 77.9 & 	12.4 / 74.2 & 	22.0 / 81.9 & 	20.2 / 82.4\\	
cmn $\rightarrow$ kor	&	22.8 / 88.6 & 	23.1 / 87.9 & 	18.0 / 84.8 & 	20.6 / 88.4 & 	22.7 / 88.0 & 	22.9 / 89.4\\	
cmn $\rightarrow$ lao	&	22.2 / 81.8 & 	21.0 / 80.9 & 	21.7 / 79.7 & 	12.6 / 71.6 & 	26.2 / 83.2 & 	25.4 / 84.6\\	
cmn $\rightarrow$ msa	&	26.7 / 87.2 & 	24.9 / 86.0 & 	24.8 / 84.6 & 	23.0 / 86.5 & 	28.6 / 86.8 & 	29.7 / 88.9\\	
cmn $\rightarrow$ mya	&	18.7 / 86.4 & 	0.2 / 27.1 & 	13.3 / 79.2 & 	12.0 / 79.5 & 	16.5 / 84.6 & 	19.5 / 87.5\\	
cmn $\rightarrow$ nld	&	26.8 / 86.8 & 	25.5 / 86.3 & 	21.5 / 82.4 & 	24.5 / 85.9 & 	26.5 / 86.3 & 	27.0 / 87.5\\	
cmn $\rightarrow$ pol	&	26.3 / 89.8 & 	25.7 / 89.1 & 	20.2 / 83.9 & 	23.6 / 88.2 & 	25.4 / 88.8 & 	25.9 / 90.3\\	
cmn $\rightarrow$ por	&	32.2 / 87.8 & 	31.9 / 87.2 & 	28.1 / 84.4 & 	31.7 / 87.5 & 	32.9 / 87.4 & 	34.0 / 87.9\\	
cmn $\rightarrow$ rus	&	31.2 / 89.5 & 	28.7 / 88.2 & 	24.8 / 85.5 & 	28.1 / 88.9 & 	28.7 / 88.8 & 	28.9 / 90.9\\	
cmn $\rightarrow$ spa	&	25.4 / 86.0 & 	24.8 / 85.7 & 	21.9 / 82.9 & 	24.6 / 85.7 & 	25.1 / 85.5 & 	26.6 / 85.7\\	
cmn $\rightarrow$ tgl	&	24.6 / 82.6 & 	24.6 / 82.3 & 	20.3 / 78.1 & 	18.4 / 79.7 & 	24.3 / 82.0 & 	26.0 / 84.5\\	
cmn $\rightarrow$ tha	&	37.3 / 87.7 & 	20.4 / 73.1 & 	27.1 / 82.6 & 	34.2 / 87.1 & 	35.5 / 86.9 & 	36.2 / 88.8\\	
cmn $\rightarrow$ tur	&	26.8 / 86.8 & 	26.0 / 86.3 & 	21.5 / 82.0 & 	24.0 / 85.4 & 	26.5 / 86.1 & 	27.3 / 88.2\\	
cmn $\rightarrow$ urd	&	21.8 / 80.3 & 	19.7 / 79.3 & 	17.4 / 74.0 & 	16.4 / 76.8 & 	21.5 / 79.7 & 	21.6 / 82.5\\	
cmn $\rightarrow$ vie	&	32.7 / 89.2 & 	32.4 / 88.9 & 	30.0 / 86.5 & 	31.6 / 88.8 & 	34.0 / 88.8 & 	34.6 / 90.4\\ 	 \midrule
Avg.	&	30.1 / 86.7 & 	27.7 / 83.0 & 	26.0 / 83.5 & 	26.4 / 84.7 & 	30.3 / 86.2 & 	31.1 / 87.7\\	
        \bottomrule
        \end{tabular}
    
    \caption{spBLEU / COMET Scores on the FLORES-200 Benchmark.}
    \label{tab:flores200}
    \end{table}

\end{document}

%% file: math_commands.tex
\usepackage{amsmath,amsfonts,bm}

\def\eqref#1{equation~\ref{#1}}

\def\1{\bm{1}}

\DeclareMathAlphabet{\mathsfit}{\encodingdefault}{\sfdefault}{m}{sl}
\SetMathAlphabet{\mathsfit}{bold}{\encodingdefault}{\sfdefault}{bx}{n}



%% file: iclr2026_conference.bib
@article{guo2025deepseek,
  title={Deepseek-r1 incentivizes reasoning in llms through reinforcement learning},
  author={Guo, Daya and Yang, Dejian and Zhang, Haowei and Song, Junxiao and Wang, Peiyi and Zhu, Qihao and Xu, Runxin and Zhang, Ruoyu and Ma, Shirong and Bi, Xiao and others},
  journal={Nature},
  volume={645},
  number={8081},
  pages={633--638},
  year={2025},
  publisher={Nature Publishing Group UK London}
}

@article{chen2024slam,
  title={SLAM-Omni: Timbre-Controllable Voice Interaction System with Single-Stage Training},
  author={Chen, Wenxi and Ma, Ziyang and Yan, Ruiqi and Liang, Yuzhe and Li, Xiquan and Xu, Ruiyang and Niu, Zhikang and Zhu, Yanqiao and Yang, Yifan and Liu, Zhanxun and others},
  journal={arXiv preprint arXiv:2412.15649},
  year={2024}
}

@article{ma2026slam,
  title={SLAM-LLM: A Modular, Open-Source Multimodal Large Language Model Framework and Best Practice for Speech, Language, Audio and Music Processing},
  author={Ma, Ziyang and Yang, Guanrou and Chen, Wenxi and Gao, Zhifu and Du, Yexing and Li, Xiquan and Zheng, Zhisheng and Zhu, Haina and Zhuo, Jianheng and Song, Zheshu and others},
  journal={IEEE Journal of Selected Topics in Signal Processing},
  year={2026},
  publisher={IEEE}
}

@inproceedings{sankar-etal-2025-towards,
    title = "Towards Building Large Scale Datasets and State-of-the-Art Automatic Speech Translation Systems for 14 {I}ndian Languages",
    author = "Sankar, Ashwin  and
      Jain, Sparsh  and
      Narasimhan, Nikhil  and
      Choudhary, Devilal  and
      Suman, Dhairya  and
      Khan, Mohammed Safi Ur Rahman  and
      Kunchukuttan, Anoop  and
      Khapra, Mitesh M  and
      Dabre, Raj",
    editor = "Che, Wanxiang  and
      Nabende, Joyce  and
      Shutova, Ekaterina  and
      Pilehvar, Mohammad Taher",
    booktitle = "Proceedings of the 63rd Annual Meeting of the Association for Computational Linguistics (Volume 1: Long Papers)",
    month = jul,
    year = "2025",
    address = "Vienna, Austria",
    publisher = "Association for Computational Linguistics",
    url = "https://aclanthology.org/2025.acl-long.1582/",
    doi = "10.18653/v1/2025.acl-long.1582",
    pages = "32945--32966",
    ISBN = "979-8-89176-251-0"
}

@article{adamw,
  title={Decoupled weight decay regularization},
  author={Loshchilov, I},
  journal={arXiv preprint arXiv:1711.05101},
  year={2017}
}

@inproceedings{kwon2023efficient,
  title={Efficient Memory Management for Large Language Model Serving with PagedAttention},
  author={Woosuk Kwon and Zhuohan Li and Siyuan Zhuang and Ying Sheng and Lianmin Zheng and Cody Hao Yu and Joseph E. Gonzalez and Hao Zhang and Ion Stoica},
  booktitle={Proceedings of the ACM SIGOPS 29th Symposium on Operating Systems Principles},
  year={2023}
}

@inproceedings{rei2020comet,
  title={COMET: A Neural Framework for MT Evaluation},
  author={Rei, Ricardo and Stewart, Craig and Farinha, Ana C and Lavie, Alon},
  booktitle={Proceedings of the 2020 Conference on Empirical Methods in Natural Language Processing (EMNLP)},
  pages={2685--2702},
  year={2020}
}

@article{tao2024survey,
  title={A survey on self-evolution of large language models},
  author={Tao, Zhengwei and Lin, Ting-En and Chen, Xiancai and Li, Hangyu and Wu, Yuchuan and Li, Yongbin and Jin, Zhi and Huang, Fei and Tao, Dacheng and Zhou, Jingren},
  journal={arXiv preprint arXiv:2404.14387},
  year={2024}
}

@article{liu2021survey,
  title={A survey on evolutionary neural architecture search},
  author={Liu, Yuqiao and Sun, Yanan and Xue, Bing and Zhang, Mengjie and Yen, Gary G and Tan, Kay Chen},
  journal={IEEE transactions on neural networks and learning systems},
  volume={34},
  number={2},
  pages={550--570},
  year={2021},
  publisher={IEEE}
}

@article{shen2024survey,
  title={A survey on multi-modal machine translation: Tasks, methods and challenges},
  author={Shen, Huangjun and Shao, Liangying and Li, Wenbo and Lan, Zhibin and Liu, Zhanyu and Su, Jinsong},
  journal={arXiv preprint arXiv:2405.12669},
  year={2024}
}

@article{cui2025multilingual,
  title={Multilingual machine translation with open large language models at practical scale: An empirical study},
  author={Cui, Menglong and Gao, Pengzhi and Liu, Wei and Luan, Jian and Wang, Bin},
  journal={arXiv preprint arXiv:2502.02481},
  year={2025}
}

@inproceedings{radford2023robust,
  title={Robust speech recognition via large-scale weak supervision},
  author={Radford, Alec and Kim, Jong Wook and Xu, Tao and Brockman, Greg and McLeavey, Christine and Sutskever, Ilya},
  booktitle={International Conference on Machine Learning},
  pages={28492--28518},
  year={2023},
  organization={PMLR}
}

@inproceedings{li2022valhalla,
  title={Valhalla: Visual hallucination for machine translation. In 2022 IEEE},
  author={Li, Yi and Panda, Rameswar and Kim, Yoon and Chen, C and Feris, Rogerio and Cox, David and Vasconcelos, Nuno},
  booktitle={CVF Conference on Computer Vision and Pattern Recognition (CVPR)},
  pages={5206--5216},
  year={2022}
}

@inproceedings{guo2023bridging,
  title={Bridging the Gap between Synthetic and Authentic Images for Multimodal Machine Translation},
  author={Guo, Wenyu and Fang, Qingkai and Yu, Dong and Feng, Yang},
  booktitle={Proceedings of the 2023 Conference on Empirical Methods in Natural Language Processing},
  pages={2863--2874},
  year={2023}
}

@article{chen2024make,
  title={Make Imagination Clearer! Stable Diffusion-based Visual Imagination for Multimodal Machine Translation},
  author={Chen, Andong and Song, Yuchen and Chen, Kehai and Yang, Muyun and Zhao, Tiejun and Zhang, Min},
  journal={arXiv preprint arXiv:2412.12627},
  year={2024}
}

@inproceedings{gao-etal-2025-multimodal,
    title = "Multimodal Machine Translation with Text-Image In-depth Questioning",
    author = "Gao, Yue  and
      Zhao, Jing  and
      Sun, Shiliang  and
      Qiao, Xiaosong  and
      Song, Tengfei  and
      Yang, Hao",
    editor = "Che, Wanxiang  and
      Nabende, Joyce  and
      Shutova, Ekaterina  and
      Pilehvar, Mohammad Taher",
    booktitle = "Findings of the Association for Computational Linguistics: ACL 2025",
    month = jul,
    year = "2025",
    address = "Vienna, Austria",
    publisher = "Association for Computational Linguistics",
    url = "https://aclanthology.org/2025.findings-acl.483/",
    doi = "10.18653/v1/2025.findings-acl.483",
    pages = "9274--9287",
    ISBN = "979-8-89176-256-5"
}

@inproceedings{chen-etal-2025-make,
    title = "Make Imagination Clearer! Stable Diffusion-based Visual Imagination for Multimodal Machine Translation",
    author = "Chen, Andong  and
      Song, Yuchen  and
      Chen, Kehai  and
      Bai, Xuefeng  and
      Yang, Muyun  and
      Nie, Liqiang  and
      Liu, Jie  and
      Zhao, Tiejun  and
      Zhang, Min",
    editor = "Che, Wanxiang  and
      Nabende, Joyce  and
      Shutova, Ekaterina  and
      Pilehvar, Mohammad Taher",
    booktitle = "Proceedings of the 63rd Annual Meeting of the Association for Computational Linguistics (Volume 1: Long Papers)",
    month = jul,
    year = "2025",
    address = "Vienna, Austria",
    publisher = "Association for Computational Linguistics",
    url = "https://aclanthology.org/2025.acl-long.1289/",
    doi = "10.18653/v1/2025.acl-long.1289",
    pages = "26567--26583",
    ISBN = "979-8-89176-251-0"
}

@article{tayir2024encoder,
  title={Encoder--Decoder Calibration for Multimodal Machine Translation},
  author={Tayir, Turghun and Li, Lin and Li, Bei and Liu, Jianquan and Lee, Kong Aik},
  journal={IEEE Transactions on Artificial Intelligence},
  volume={5},
  number={8},
  pages={3965--3973},
  year={2024},
  publisher={IEEE}
}

@inproceedings{li2021vision,
  title={Vision Matters When It Should: Sanity Checking Multimodal Machine Translation Models},
  author={Li, Jiaoda and Ataman, Duygu and Sennrich, Rico},
  booktitle={2021 Conference on Empirical Methods in Natural Language Processing},
  pages={8556--8562},
  year={2021},
  organization={Association for Computational Linguistics}
}

@inproceedings{joshi2020state,
  title={The State and Fate of Linguistic Diversity and Inclusion in the NLP World},
  author={Joshi, Pratik and Santy, Sebastin and Budhiraja, Amar and Bali, Kalika and Choudhury, Monojit},
  booktitle={Proceedings of the 58th Annual Meeting of the Association for Computational Linguistics},
  pages={6282},
  year={2020},
  organization={Association for Computational Linguistics}
}

@article{DBLP:journals/taslp/ZhaoKKC22,
  author       = {Yuting Zhao and
                  Mamoru Komachi and
                  Tomoyuki Kajiwara and
                  Chenhui Chu},
  title        = {Word-Region Alignment-Guided Multimodal Neural Machine Translation},
  journal      = {{IEEE} {ACM} Trans. Audio Speech Lang. Process.},
  volume       = {30},
  pages        = {244--259},
  year         = {2022},
  url          = {https://doi.org/10.1109/TASLP.2021.3138719},
  doi          = {10.1109/TASLP.2021.3138719},
  bibsource    = {dblp computer science bibliography, https://dblp.org}
}

@inproceedings{DBLP:conf/ijcnlp/ElliottK17,
  author    = {Desmond Elliott and
               {\'{A}}kos K{\'{a}}d{\'{a}}r},
  editor    = {Greg Kondrak and
               Taro Watanabe},
  title     = {Imagination Improves Multimodal Translation},
  booktitle = {Proceedings of the Eighth International Joint Conference on Natural
               Language Processing, {IJCNLP} 2017, Taipei, Taiwan, November 27 -
               December 1, 2017 - Volume 1: Long Papers},
  pages     = {130--141},
  publisher = {Asian Federation of Natural Language Processing},
  year      = {2017},
  url       = {https://aclanthology.org/I17-1014/},
  bibsource = {dblp computer science bibliography, https://dblp.org}
}

@inproceedings{DBLP:conf/acl/ZhuSCHWW23,
  author       = {Yaoming Zhu and
                  Zewei Sun and
                  Shanbo Cheng and
                  Luyang Huang and
                  Liwei Wu and
                  Mingxuan Wang},
  editor       = {Anna Rogers and
                  Jordan L. Boyd{-}Graber and
                  Naoaki Okazaki},
  title        = {Beyond Triplet: Leveraging the Most Data for Multimodal Machine Translation},
  booktitle    = {Findings of the Association for Computational Linguistics: {ACL} 2023,
                  Toronto, Canada, July 9-14, 2023},
  pages        = {2679--2697},
  publisher    = {Association for Computational Linguistics},
  year         = {2023},
  url          = {https://doi.org/10.18653/v1/2023.findings-acl.168},
  doi          = {10.18653/V1/2023.FINDINGS-ACL.168},
  bibsource    = {dblp computer science bibliography, https://dblp.org}
}

@inproceedings{DBLP:conf/mm/0003CJLXH21,
  author    = {Yuqing Song and
               Shizhe Chen and
               Qin Jin and
               Wei Luo and
               Jun Xie and
               Fei Huang},
  editor    = {Heng Tao Shen and
               Yueting Zhuang and
               John R. Smith and
               Yang Yang and
               Pablo C{\'{e}}sar and
               Florian Metze and
               Balakrishnan Prabhakaran},
  title     = {Product-oriented Machine Translation with Cross-modal Cross-lingual
               Pre-training},
  booktitle = {{MM} '21: {ACM} Multimedia Conference, Virtual Event, China, October
               20 - 24, 2021},
  pages     = {2843--2852},
  publisher = {{ACM}},
  year      = {2021},
  url       = {https://doi.org/10.1145/3474085.3475303},
  doi       = {10.1145/3474085.3475303},
  bibsource = {dblp computer science bibliography, https://dblp.org}
}

@inproceedings{gao2025multimodal,
  title     = {Multimodal Machine Translation with Text-Image In-depth Questioning},
  author    = {Gao, Yue and Zhao, Jing and Sun, Shiliang and Qiao, Xiaosong and Song, Tengfei and Yang, Hao},
  booktitle = {Proceedings of the 63rd Annual Meeting of the Association for Computational Linguistics (ACL 2025)},
  year      = {2025},
}

@article{DBLP:journals/tmm/ChenYYL23,
  author       = {Zhuo Chen and
                  Fei Yin and
                  Qing Yang and
                  Cheng{-}Lin Liu},
  title        = {Cross-Lingual Text Image Recognition via Multi-Hierarchy Cross-Modal
                  Mimic},
  journal      = {{IEEE} Trans. Multim.},
  volume       = {25},
  pages        = {4830--4841},
  year         = {2023},
}

@article{team2025gemma,
  title={Gemma 3 technical report},
  author={Team, Gemma and Kamath, Aishwarya and Ferret, Johan and Pathak, Shreya and Vieillard, Nino and Merhej, Ramona and Perrin, Sarah and Matejovicova, Tatiana and Ram{\'e}, Alexandre and Rivi{\`e}re, Morgane and others},
  journal={arXiv preprint arXiv:2503.19786},
  year={2025}
}

@inproceedings{DBLP:conf/icpr/MaZTHWZ022,
  author    = {Cong Ma and
               Yaping Zhang and
               Mei Tu and
               Xu Han and
               Linghui Wu and
               Yang Zhao and
               Yu Zhou},
  title     = {Improving End-to-End Text Image Translation From the Auxiliary Text
               Translation Task},
  booktitle = {26th International Conference on Pattern Recognition, {ICPR} 2022,
               Montreal, QC, Canada, August 21-25, 2022},
  pages     = {1664--1670},
  publisher = {{IEEE}},
  year      = {2022},
  url       = {https://doi.org/10.1109/ICPR56361.2022.9956695},
  doi       = {10.1109/ICPR56361.2022.9956695},
  bibsource = {dblp computer science bibliography, https://dblp.org}
}

@inproceedings{DBLP:conf/acl/LanYLZ0WHS23,
  author       = {Zhibin Lan and
                  Jiawei Yu and
                  Xiang Li and
                  Wen Zhang and
                  Jian Luan and
                  Bin Wang and
                  Degen Huang and
                  Jinsong Su},
  editor       = {Anna Rogers and
                  Jordan L. Boyd{-}Graber and
                  Naoaki Okazaki},
  title        = {Exploring Better Text Image Translation with Multimodal Codebook},
  booktitle    = {Proceedings of the 61st Annual Meeting of the Association for Computational
                  Linguistics (Volume 1: Long Papers), {ACL} 2023, Toronto, Canada,
                  July 9-14, 2023},
  pages        = {3479--3491},
  publisher    = {Association for Computational Linguistics},
  year         = {2023},
  url          = {https://doi.org/10.18653/v1/2023.acl-long.192},
  doi          = {10.18653/V1/2023.ACL-LONG.192},
  bibsource    = {dblp computer science bibliography, https://dblp.org}
}

@inproceedings{DBLP:conf/acl/LiangMXCZ22,
  author    = {Yunlong Liang and
               Fandong Meng and
               Jinan Xu and
               Yufeng Chen and
               Jie Zhou},
  editor    = {Smaranda Muresan and
               Preslav Nakov and
               Aline Villavicencio},
  title     = {{MSCTD:} {A} Multimodal Sentiment Chat Translation Dataset},
  booktitle = {Proceedings of the 60th Annual Meeting of the Association for Computational
               Linguistics (Volume 1: Long Papers), {ACL} 2022, Dublin, Ireland,
               May 22-27, 2022},
  pages     = {2601--2613},
  publisher = {Association for Computational Linguistics},
  year      = {2022},
  url       = {https://doi.org/10.18653/v1/2022.acl-long.186},
  doi       = {10.18653/v1/2022.acl-long.186},
  bibsource = {dblp computer science bibliography, https://dblp.org}
}

@inproceedings{rombach2022high,
  title={High-resolution image synthesis with latent diffusion models},
  author={Rombach, Robin and Blattmann, Andreas and Lorenz, Dominik and Esser, Patrick and Ommer, Bj{\"o}rn},
  booktitle={Proceedings of the IEEE/CVF conference on computer vision and pattern recognition},
  pages={10684--10695},
  year={2022}
}

@article{nllb2022,
  author    = {NLLB Team and Marta R. Costa-jussà and James Cross and Onur Çelebi and Maha Elbayad and Kenneth Heafield and Kevin Heffernan and Elahe Kalbassi and Janice Lam and Daniel Licht and Jean Maillard and Anna Sun and Skyler Wang and Guillaume Wenzek and Al Youngblood and Bapi Akula and Loic Barrault and Gabriel Mejia Gonzalez and Prangthip Hansanti and John Hoffman and Semarley Jarrett and Kaushik Ram Sadagopan and Dirk Rowe and Shannon Spruit and Chau Tran and Pierre Andrews and Necip Fazil Ayan and Shruti Bhosale and Sergey Edunov and Angela Fan and Cynthia Gao and Vedanuj Goswami and Francisco Guzmán and Philipp Koehn and Alexandre Mourachko and Christophe Ropers and Safiyyah Saleem and Holger Schwenk and Jeff Wang},
  title     = {No Language Left Behind: Scaling Human-Centered Machine Translation},
  year      = {2022}
}

@inproceedings{DBLP:conf/icml/0008LSH23,
  author       = {Junnan Li and
                  Dongxu Li and
                  Silvio Savarese and
                  Steven C. H. Hoi},
  editor       = {Andreas Krause and
                  Emma Brunskill and
                  Kyunghyun Cho and
                  Barbara Engelhardt and
                  Sivan Sabato and
                  Jonathan Scarlett},
  title        = {{BLIP-2:} Bootstrapping Language-Image Pre-training with Frozen Image
                  Encoders and Large Language Models},
  booktitle    = {International Conference on Machine Learning, {ICML} 2023, 23-29 July
                  2023, Honolulu, Hawaii, {USA}},
  series       = {Proceedings of Machine Learning Research},
  volume       = {202},
  pages        = {19730--19742},
  publisher    = {{PMLR}},
  year         = {2023},
  url          = {https://proceedings.mlr.press/v202/li23q.html},
  bibsource    = {dblp computer science bibliography, https://dblp.org}
}

@inproceedings{DBLP:conf/emnlp/GuoLHYL0C22,
  author    = {Hongcheng Guo and
               Jiaheng Liu and
               Haoyang Huang and
               Jian Yang and
               Zhoujun Li and
               Dongdong Zhang and
               Zheng Cui},
  editor    = {Yoav Goldberg and
               Zornitsa Kozareva and
               Yue Zhang},
  title     = {{LVP-M3:} Language-aware Visual Prompt for Multilingual Multimodal
               Machine Translation},
  booktitle = {Proceedings of the 2022 Conference on Empirical Methods in Natural
               Language Processing, {EMNLP} 2022, Abu Dhabi, United Arab Emirates,
               December 7-11, 2022},
  pages     = {2862--2872},
  publisher = {Association for Computational Linguistics},
  year      = {2022},
  url       = {https://aclanthology.org/2022.emnlp-main.184},
  bibsource = {dblp computer science bibliography, https://dblp.org}
}

@article{Grubinger_2006,  
 title={The IAPR Benchmark: A New Evaluation Resource for Visual Information Systems}, 
 journal={Language Resources and Evaluation}, 
 author={Grubinger, M.}, 
 year={2006}, 
 month={Jan}, 
 language={en-US} 
 }

@inproceedings{DBLP:conf/acl/ElliottFSS16,
  author       = {Desmond Elliott and
                  Stella Frank and
                  Khalil Sima'an and
                  Lucia Specia},
  title        = {Multi30K: Multilingual English-German Image Descriptions},
  booktitle    = {Proceedings of the 5th Workshop on Vision and Language, hosted by
                  the 54th Annual Meeting of the Association for Computational Linguistics,
                  VL@ACL 2016, August 12, Berlin, Germany},
  publisher    = {The Association for Computer Linguistics},
  year         = {2016},
  url          = {https://doi.org/10.18653/v1/w16-3210},
  doi          = {10.18653/v1/w16-3210},
  bibsource    = {dblp computer science bibliography, https://dblp.org}
}

@inproceedings{DBLP:conf/acl/SikasoteMAA23,
  author       = {Claytone Sikasote and
                  Eunice Mukonde and
                  Md Mahfuz Ibn Alam and
                  Antonios Anastasopoulos},
  editor       = {Anna Rogers and
                  Jordan L. Boyd{-}Graber and
                  Naoaki Okazaki},
  title        = {{BIG-C:} a Multimodal Multi-Purpose Dataset for Bemba},
  booktitle    = {Proceedings of the 61st Annual Meeting of the Association for Computational
                  Linguistics (Volume 1: Long Papers), {ACL} 2023, Toronto, Canada,
                  July 9-14, 2023},
  pages        = {2062--2078},
  publisher    = {Association for Computational Linguistics},
  year         = {2023},
  url          = {https://doi.org/10.18653/v1/2023.acl-long.115},
  doi          = {10.18653/V1/2023.ACL-LONG.115},
  bibsource    = {dblp computer science bibliography, https://dblp.org}
}

@inproceedings{DBLP:conf/acl/ParidaAMBKAKSBK23,
  author       = {Shantipriya Parida and
                  Idris Abdulmumin and
                  Shamsuddeen Hassan Muhammad and
                  Aneesh Bose and
                  Guneet Singh Kohli and
                  Ibrahim Said Ahmad and
                  Ketan Kotwal and
                  Sayan Deb Sarkar and
                  Ondrej Bojar and
                  Habeebah A. Kakudi},
  editor       = {Anna Rogers and
                  Jordan L. Boyd{-}Graber and
                  Naoaki Okazaki},
  title        = {HaVQA: {A} Dataset for Visual Question Answering and Multimodal Research
                  in Hausa Language},
  booktitle    = {Findings of the Association for Computational Linguistics: {ACL} 2023,
                  Toronto, Canada, July 9-14, 2023},
  pages        = {10162--10183},
  publisher    = {Association for Computational Linguistics},
  year         = {2023},
  url          = {https://doi.org/10.18653/v1/2023.findings-acl.646},
  doi          = {10.18653/V1/2023.FINDINGS-ACL.646},
  bibsource    = {dblp computer science bibliography, https://dblp.org}
}

@inproceedings{DBLP:conf/naacl/GellaEK19,
  author       = {Spandana Gella and
                  Desmond Elliott and
                  Frank Keller},
  editor       = {Jill Burstein and
                  Christy Doran and
                  Thamar Solorio},
  title        = {Cross-lingual Visual Verb Sense Disambiguation},
  booktitle    = {Proceedings of the 2019 Conference of the North American Chapter of
                  the Association for Computational Linguistics: Human Language Technologies,
                  {NAACL-HLT} 2019, Minneapolis, MN, USA, June 2-7, 2019, Volume 1 (Long
                  and Short Papers)},
  pages        = {1998--2004},
  publisher    = {Association for Computational Linguistics},
  year         = {2019},
  url          = {https://doi.org/10.18653/v1/n19-1200},
  doi          = {10.18653/v1/n19-1200},
  bibsource    = {dblp computer science bibliography, https://dblp.org}
}

@inproceedings{DBLP:conf/lrec/LalaS18,
  author       = {Chiraag Lala and
                  Lucia Specia},
  editor       = {Nicoletta Calzolari and
                  Khalid Choukri and
                  Christopher Cieri and
                  Thierry Declerck and
                  Sara Goggi and
                  K{\^{o}}iti Hasida and
                  Hitoshi Isahara and
                  Bente Maegaard and
                  Joseph Mariani and
                  H{\'{e}}l{\`{e}}ne Mazo and
                  Asunci{\'{o}}n Moreno and
                  Jan Odijk and
                  Stelios Piperidis and
                  Takenobu Tokunaga},
  title        = {Multimodal Lexical Translation},
  booktitle    = {Proceedings of the Eleventh International Conference on Language Resources
                  and Evaluation, {LREC} 2018, Miyazaki, Japan, May 7-12, 2018},
  publisher    = {European Language Resources Association {(ELRA)}},
  year         = {2018},
  url          = {http://www.lrec-conf.org/proceedings/lrec2018/summaries/629.html},
  bibsource    = {dblp computer science bibliography, https://dblp.org}
}

@inproceedings{yu2024connecting,
  title={Connecting speech encoder and large language model for asr},
  author={Yu, Wenyi and Tang, Changli and Sun, Guangzhi and Chen, Xianzhao and Tan, Tian and Li, Wei and Lu, Lu and Ma, Zejun and Zhang, Chao},
  booktitle={ICASSP 2024-2024 IEEE International Conference on Acoustics, Speech and Signal Processing (ICASSP)},
  pages={12637--12641},
  year={2024},
  organization={IEEE}
}

@article{koluguri2025granary,
  title={Granary: Speech Recognition and Translation Dataset in 25 European Languages},
  author={Koluguri, Nithin Rao and Sekoyan, Monica and Zelenfroynd, George and Meister, Sasha and Ding, Shuoyang and Kostandian, Sofia and Huang, He and Karpov, Nikolay and Balam, Jagadeesh and Lavrukhin, Vitaly and others},
  journal={arXiv preprint arXiv:2505.13404},
  year={2025}
}

@article{du2025ccfqa,
  title={CCFQA: A Benchmark for Cross-Lingual and Cross-Modal Speech and Text Factuality Evaluation},
  author={Du, Yexing and Liu, Kaiyuan and Pan, Youcheng and Chu, Zheng and Yang, Bo and Feng, Xiaocheng and Xiang, Yang and Liu, Ming},
  journal={arXiv preprint arXiv:2508.07295},
  year={2025}
}

@online{qwen3-blog-2024,
  author       = {Qwen Team},
  title        = {Qwen3: Think Deeper, Act Faster},
  year         = {2024},
  url          = {https://qwenlm.github.io/zh/blog/qwen3/},
  urldate      = {2024-06-20},
  language     = {zh},
  organization = {QwenLM}
}

@online{qwen3-next-blog-2025,
  author       = {Qwen Team},
  title        = {Qwen3-Next: More Power and Less Cost},
  year         = {2025},
  url          = {https://qwen.ai/blog?id=4074cca80393150c248e508aa62983f9cb7d27cd&from=research.latest-advancements-list},
  urldate      = {2025-09},
  language     = {zh},
  organization = {QwenLM}
}

@article{deutsch2025wmt24++,
  title={Wmt24++: Expanding the language coverage of wmt24 to 55 languages \& dialects},
  author={Deutsch, Daniel and Briakou, Eleftheria and Caswell, Isaac and Finkelstein, Mara and Galor, Rebecca and Juraska, Juraj and Kovacs, Geza and Lui, Alison and Rei, Ricardo and Riesa, Jason and others},
  journal={arXiv preprint arXiv:2502.12404},
  year={2025}
}

@article{fleurs2022arxiv,
  title = {FLEURS: Few-shot Learning Evaluation of Universal Representations of Speech},
  author = {Conneau, Alexis and Ma, Min and Khanuja, Simran and Zhang, Yu and Axelrod, Vera and Dalmia, Siddharth and Riesa, Jason and Rivera, Clara and Bapna, Ankur},
  journal={arXiv preprint arXiv:2205.12446},
  url = {https://arxiv.org/abs/2205.12446},
  year = {2022},
}

@inproceedings{di2019must,
  title={Must-c: a multilingual speech translation corpus},
  author={Di Gangi, Mattia A and Cattoni, Roldano and Bentivogli, Luisa and Negri, Matteo and Turchi, Marco},
  booktitle={Proceedings of the 2019 Conference of the North American Chapter of the Association for Computational Linguistics: Human Language Technologies, Volume 1 (Long and Short Papers)},
  pages={2012--2017},
  year={2019},
  organization={Association for Computational Linguistics}
}

@inproceedings{iranzo2020europarl,
  title={Europarl-st: A multilingual corpus for speech translation of parliamentary debates},
  author={Iranzo-S{\'a}nchez, Javier and Silvestre-Cerda, Joan Albert and Jorge, Javier and Rosell{\'o}, Nahuel and Gim{\'e}nez, Adria and Sanchis, Albert and Civera, Jorge and Juan, Alfons},
  booktitle={ICASSP 2020-2020 IEEE International Conference on Acoustics, Speech and Signal Processing (ICASSP)},
  pages={8229--8233},
  year={2020},
  organization={IEEE}
}

@inproceedings{li2023blip,
  title={Blip-2: Bootstrapping language-image pre-training with frozen image encoders and large language models},
  author={Li, Junnan and Li, Dongxu and Savarese, Silvio and Hoi, Steven},
  booktitle={International conference on machine learning},
  pages={19730--19742},
  year={2023},
  organization={PMLR}
}

@article{wang2020covost,
  title={Covost 2 and massively multilingual speech-to-text translation},
  author={Wang, Changhan and Wu, Anne and Pino, Juan},
  journal={arXiv preprint arXiv:2007.10310},
  year={2020}
}

@inproceedings{post-2018-call,
    title = "A Call for Clarity in Reporting {BLEU} Scores",
    author = "Post, Matt",
    booktitle = "Proceedings of the Third Conference on Machine Translation: Research Papers",
    month = oct,
    year = "2018",
    address = "Belgium, Brussels",
    publisher = "Association for Computational Linguistics",
    url = "https://www.aclweb.org/anthology/W18-6319",
    pages = "186--191",
}

@inproceedings{chi2025role,
  title={The Role of Prosody in Spoken Question Answering},
  author={Chi, Jie and de Seyssel, Maureen and Schluter, Natalie},
  booktitle={Findings of the Association for Computational Linguistics: NAACL 2025},
  pages={8468--8479},
  year={2025}
}

@article{du2024cosyvoice,
  title={Cosyvoice: A scalable multilingual zero-shot text-to-speech synthesizer based on supervised semantic tokens},
  author={Du, Zhihao and Chen, Qian and Zhang, Shiliang and Hu, Kai and Lu, Heng and Yang, Yexin and Hu, Hangrui and Zheng, Siqi and Gu, Yue and Ma, Ziyang and others},
  journal={arXiv preprint arXiv:2407.05407},
  year={2024}
}

@inproceedings{cheng-etal-2024-soul,
    title = "Soul-Mix: Enhancing Multimodal Machine Translation with Manifold Mixup",
    author = "Cheng, Xuxin  and
      Yao, Ziyu  and
      Xin, Yifei  and
      An, Hao  and
      Li, Hongxiang  and
      Li, Yaowei  and
      Zou, Yuexian",
    editor = "Ku, Lun-Wei  and
      Martins, Andre  and
      Srikumar, Vivek",
    booktitle = "Proceedings of the 62nd Annual Meeting of the Association for Computational Linguistics (Volume 1: Long Papers)",
    month = aug,
    year = "2024",
    address = "Bangkok, Thailand",
    publisher = "Association for Computational Linguistics",
    url = "https://aclanthology.org/2024.acl-long.608",
    doi = "10.18653/v1/2024.acl-long.608",
    pages = "11283--11294",
}

@article{DBLP:journals/talip/TayirL24,
  author       = {Turghun Tayir and
                  Lin Li},
  title        = {Unsupervised Multimodal Machine Translation for Low-resource Distant
                  Language Pairs},
  journal      = {{ACM} Trans. Asian Low Resour. Lang. Inf. Process.},
  volume       = {23},
  number       = {4},
  pages        = {55},
  year         = {2024},
  url          = {https://doi.org/10.1145/3652161},
  doi          = {10.1145/3652161},
  bibsource    = {dblp computer science bibliography, https://dblp.org}
}

@inproceedings{DBLP:conf/acl/HitschlerSR16,
  author       = {Julian Hitschler and
                  Shigehiko Schamoni and
                  Stefan Riezler},
  title        = {Multimodal Pivots for Image Caption Translation},
  booktitle    = {Proceedings of the 54th Annual Meeting of the Association for Computational
                  Linguistics, {ACL} 2016, August 7-12, 2016, Berlin, Germany, Volume
                  1: Long Papers},
  publisher    = {The Association for Computer Linguistics},
  year         = {2016},
  url          = {https://doi.org/10.18653/v1/p16-1227},
  doi          = {10.18653/V1/P16-1227},
  bibsource    = {dblp computer science bibliography, https://dblp.org}
}

@inproceedings{DBLP:conf/iclr/DongHPQGYZSZWK024,
  author       = {Runpei Dong and
                  Chunrui Han and
                  Yuang Peng and
                  Zekun Qi and
                  Zheng Ge and
                  Jinrong Yang and
                  Liang Zhao and
                  Jianjian Sun and
                  Hongyu Zhou and
                  Haoran Wei and
                  Xiangwen Kong and
                  Xiangyu Zhang and
                  Kaisheng Ma and
                  Li Yi},
  title        = {DreamLLM: Synergistic Multimodal Comprehension and Creation},
  booktitle    = {The Twelfth International Conference on Learning Representations,
                  {ICLR} 2024, Vienna, Austria, May 7-11, 2024},
  publisher    = {OpenReview.net},
  year         = {2024},
  url          = {https://openreview.net/forum?id=y01KGvd9Bw},
  bibsource    = {dblp computer science bibliography, https://dblp.org}
}

@inproceedings{du-etal-2025-making,
    title = "Making {LLM}s Better Many-to-Many Speech-to-Text Translators with Curriculum Learning",
    author = "Du, Yexing  and
      Pan, Youcheng  and
      Ma, Ziyang  and
      Yang, Bo  and
      Yang, Yifan  and
      Deng, Keqi  and
      Chen, Xie  and
      Xiang, Yang  and
      Liu, Ming  and
      Qin, Bing",
    editor = "Che, Wanxiang  and
      Nabende, Joyce  and
      Shutova, Ekaterina  and
      Pilehvar, Mohammad Taher",
    booktitle = "Proceedings of the 63rd Annual Meeting of the Association for Computational Linguistics (Volume 1: Long Papers)",
    month = jul,
    year = "2025",
    address = "Vienna, Austria",
    publisher = "Association for Computational Linguistics",
    url = "https://aclanthology.org/2025.acl-long.610/",
    doi = "10.18653/v1/2025.acl-long.610",
    pages = "12466--12478",
    ISBN = "979-8-89176-251-0"
}

@misc{du2025mcatscalingmanytomanyspeechtotext,
      title={MCAT: Scaling Many-to-Many Speech-to-Text Translation with MLLMs to 70 Languages}, 
      author={Yexing Du and Kaiyuan Liu and Youcheng Pan and Bo Yang and Keqi Deng and Xie Chen and Yang Xiang and Ming Liu and Bin Qin and YaoWei Wang},
      year={2025},
      eprint={2512.01512},
      archivePrefix={arXiv},
      primaryClass={cs.CL},
      url={https://arxiv.org/abs/2512.01512}, 
}
